\documentclass{article}
\usepackage{graphicx} 
\usepackage{amsmath}
\usepackage{amsfonts}
\usepackage{float}
\usepackage{url}
\usepackage[toc,page]{appendix}
\usepackage{tabularx}
\usepackage[margin=1in]{geometry}
\usepackage{multicol}
\usepackage{array}
\usepackage{color}
\usepackage{booktabs}
\usepackage[normalem]{ulem}

\usepackage{silence}
\title{
Simulation as Supervision: Mechanistic Pretraining for Scientific Discovery}
\author{Carson Dudley, Reiden Magdaleno, Christopher Harding, Marisa Eisenberg}
\date{\today}

\begin{document}

\maketitle

\section{Abstract}

Scientific modeling faces a tradeoff between the interpretability of mechanistic theory and the predictive power of machine learning. While existing hybrid approaches have made progress by incorporating domain knowledge into machine learning methods as functional constraints, they can be limited by a reliance on precise mathematical specifications. When the underlying equations are partially unknown or misspecified, enforcing rigid constraints can introduce bias and hinder a model's ability to learn from data. We introduce Simulation-Grounded Neural Networks (SGNNs), a framework that incorporates scientific theory by using mechanistic simulations as training data for neural networks. By pretraining on diverse synthetic corpora that span multiple model structures and realistic observational noise, SGNNs internalize the underlying dynamics of a system as a structural prior.

We evaluated SGNNs across multiple disciplines, including epidemiology, ecology, social science, and chemistry. In forecasting tasks, SGNNs outperformed both standard data-driven baselines and physics-constrained hybrid models. They nearly tripled the forecasting skill of the average CDC models in COVID-19 mortality forecasts and accurately forecasted high-dimensional ecological systems. SGNNs demonstrated robustness to model misspecification, performing well even when trained on data with incorrect assumptions. Our framework also introduces back-to-simulation attribution, a method for mechanistic interpretability that explains real-world dynamics by identifying their most similar counterparts within the simulated corpus. By unifying these techniques into a single framework, we demonstrate that diverse mechanistic simulations can serve as effective training data for robust scientific inference.

\section{Introduction}

Scientific modeling is often divided between two approaches: theory-driven models, which encode domain knowledge into mathematical equations, and data-driven models, which learn flexible patterns from large datasets. Each has fundamental limitations. Mechanistic models are interpretable and grounded in scientific theory, but often break down in practice due to noise, unobserved variables, or incorrect structural assumptions. Neural networks can capture complex relationships but require large labeled datasets, lack scientific grounding, and often learn spurious patterns that fail when applied to new data. There remains a critical need for modeling frameworks that are both scientifically grounded and capable of robust generalization from imperfect real-world data.

Recent work has sought to bridge this divide by incorporating mechanistic knowledge into machine learning. These approaches can be understood through the three core components of machine learning problems: the data, the model architecture, and the optimization procedure. Physics-Informed Neural Networks (PINNs) incorporate governing equations into the loss function, constraining optimization to satisfy known physical laws \cite{pinns, rodriguez2022einns}. Universal Differential Equations (UDEs) incorporate domain knowledge into the model architecture by using neural networks to represent unknown terms within a larger system of differential equations \cite{udes}. These methods perform well when equations are known and all relevant variables are observable, but can introduce bias or cause the learning process to fail when mechanistic models are incomplete or structurally misspecified.

Another approach is to incorporate mechanistic knowledge into the training data itself by using simulations to generate synthetic datasets for neural network training. This strategy has emerged across several disciplines: in physics, simulation-trained networks often serve as fast surrogates for computationally expensive models \cite{well, abm_surrogate}, while in epidemiology and social science, they can address data scarcity issues by generating synthetic trianing data \cite{defsi}. However, many of these efforts rely on a single, fixed simulator, making them brittle in the same sense as methods that directly constrain neural network outputs. While training on a single model structure is effective in domains like physics where governing equations are well-understood, it is less effective in fields where the data-generating process is fundamentally uncertain. In an emerging pandemic, for instance, a model restricted to a single mechanistic hypothesis cannot account for the vast range of plausible transmission dynamics, causing the neural network to memorize the specific biases of that model. Despite promising empirical results in isolated contexts, these approaches lack a unified framework for achieving robustness through mechanistic diversity.

We introduce Simulation-Grounded Neural Networks (SGNNs) as a general framework for training neural networks on ensembles of mechanistic simulations. While simulation-grounded learning has been applied in a range of different contexts, our goal here is to provide a unified treatment and evaluation across domains. By pretraining on synthetic corpora spanning diverse model structures, parameter regimes, and realistic observational artifacts, the model internalizes the global geometry of physically plausible dynamics. This allows the neural network to inherit scientific structure from the underlying models without being explicitly constrained by a single, potentially misspecified governing model. Our framework is architecture-agnostic and maintains flexibility across diverse scientific domains.

SGNNs are aligned with recent work demonstrating that large neural networks can acquire general-purpose inference strategies through training on synthetic data \cite{pfn, tabpfn, timepfn}. However, while previous models in this space rely on generic statistical priors to generate training data---such as Gaussian processes or broad distributions of time series---SGNNs use domain-specific scientific simulations to generate a training corpus. This shift from statistical to mechanistic supervision ensures that the model internalizes the causal structure and constraints of the physical world. Consequently, SGNNs yield models that are not only performant but also mechanistically interpretable and inherently aligned with scientific theory.

This approach directly addresses a core limitation of machine learning in science: the inability to infer unobservable quantities. Scientific parameters like the basic reproduction number in an infectious disease outbreak or ecological carrying capacity cannot be measured directly. Traditional methods estimate them by fitting mechanistic models to observed trajectories, a process sensitive to noise, delays, and structural mismatch. Simulation-based inference (SBI) \cite{sbi1, sbi2, sbi3} addresses this by using simulation-grounded training to learn posterior distributions over parameters, providing a principled approach to Bayesian parameter inference. However, full posterior inference is not always necessary: simulation-grounded learning can also enable simpler approaches such as direct quantile regression on parameters, which can achieve accurate inference with reduced computational complexity.

SGNNs also enable back-to-simulation attribution, a form of mechanistic interpretability. After training, the model retrieves training simulations that it considers most similar to real-world inputs, revealing which mechanistic regimes the model associates with observed data. Because retrieved simulations have known ground-truth parameters, users can directly inspect the model's implicit beliefs about underlying processes. Unlike post hoc feature attribution methods such as SHAP \cite{shap} or LIME \cite{lime}, back-to-simulation attribution provides process-level explanations: not just which input features mattered, but which scientific mechanisms the model believes are active.

We evaluate the SGNN framework across four distinct scientific domains: epidemiology, ecology, chemistry, and social science. Our experiments cover a broad range of tasks, including time-series forecasting, parameter inference, yield regression, and network classification. In ecological forecasting, SGNNs maintain robust accuracy in relatively high-dimensional settings (up to 16 species) where task-specific neural networks often fail to generalize. In chemical synthesis, the model reduces residual variance in reaction yield prediction by one-third compared to standard data-driven baselines. For infectious disease forecasting, SGNNs pretrained on synthetic data demonstrate competitive performance against real-world benchmarks, nearly tripling the median skill of the CDC Forecast Hub during the early COVID-19 pandemic and outperforming specialized models on vector-borne dengue outbreaks by 24\%. Finally, in social network analysis, SGNNs accurately identify the origin of diffusion cascades under conditions of partial observability. Together, these results suggest that mechanistic simulations can provide a robust foundation for inference across disparate scientific tasks. \footnote{Code and data available at \url{https://github.com/carsondudley1/SGNNs}.}

\section{Simulation-Grounded Neural Network (SGNN) Framework}

\begin{figure}[H]
    \centering
    \includegraphics[width=\textwidth]{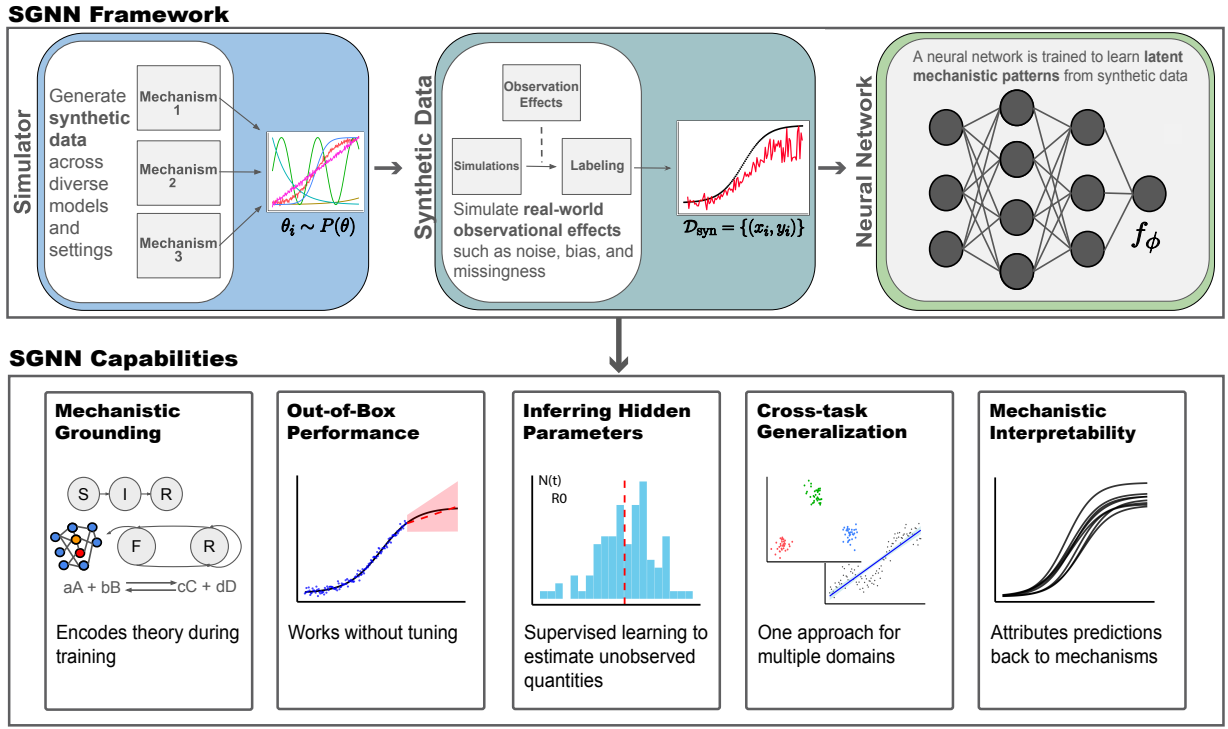}
    \caption{\textbf{Simulation-Grounded Neural Networks (SGNNs) use mechanistic simulations as synthetic supervision for scientific modeling.} \textbf{Top:} SGNNs are trained on synthetic datasets generated by mechanistic simulators spanning diverse model structures and parameter regimes. Real-world observational artifacts---such as noise, bias, and missingness---are explicitly simulated to mimic real-world data. Neural networks are trained on this synthetic corpus to learn latent mechanistic patterns, enabling generalization to real-world data without retraining. \textbf{Bottom:} SGNNs unify the strengths of mechanistic and data-driven models. They encode scientific theory during training, generalize out-of-the-box without tuning, enable supervised learning for unobservable scientific quantities, support cross-task modeling across domains, and provide process-level interpretability through back-to-simulation attribution.}
    \label{fig:sgnn_pipeline}
\end{figure}

\subsection{Conceptual Framework}

SGNNs train neural networks on synthetic data generated by mechanistic simulators. Let $S(\theta)$ denote a stochastic simulator parameterized by $\theta$, which generates trajectories or time series $x_i = S(\theta_i)$. $\theta$ sets the parameter values and decides which of multiple mechanisms will be turned on for each simulation. To capture a wide range of system behaviors, we sample $\theta_i \sim P(\theta)$ from a task-specific distribution that spans multiple model classes, parameter regimes, and observational settings. Importantly, simulations are passed through an observation model that adds realistic artifacts such as noise, censoring, delays, or reporting bias.

Each simulation yields one or more training examples $x_i$ paired with targets $y_i = T(\theta_i)$, where $T(\cdot)$ defines the labeling function (e.g., future trajectories, final states, or latent scientific quantities). The result is a synthetic dataset $\mathcal{D}_{\text{syn}} = \{(x_i, y_i)\}$ that reflects both mechanistic diversity and realistic surveillance scenarios.

A neural network $f_\phi$ is then trained via supervised learning to predict $y_i$ from $x_i$. Because each input is paired with exact labels, including latent scientific quantities, and the training set spans a wide range of mechanistic regimes and observational conditions, our aim is that the model learns to internalize structure that generalizes beyond any single model or parameter setting. SGNNs can be built using a range of possible architectures, so long as they have sufficient capacity. In practice, we use domain-specific backbones such as temporal transformers, graph networks, or hybrid models (Appendix~\ref{appendix:architecture}).

This simulation-grounded approach supports direct supervision for both observable and unobservable targets, and---with a sufficient mechanistic model training set---enables out-of-the-box generalization to real-world data.

\subsection{Key Capabilities}

We aim for Simulation-Grounded Neural Networks (SGNNs) to address several core limitations of both mechanistic and data-driven modeling approaches. Our goal for the framework is to provide five key capabilities:

\begin{itemize}
\item \textbf{Mechanistic grounding.} SGNNs learn from diverse, theory-driven simulations that encode plausible system behaviors, enabling the SGNN to internalize scientific structure.

\item \textbf{Robustness to misspecification.} By training on an ensemble of models spanning multiple structures, parameter regimes, and noise processes, SGNNs should develop flexible representations that generalize across regimes, even when real-world dynamics deviate from any one simulator.

\item \textbf{Supervised learning for unobservable targets.} SGNNs enable direct supervision on latent scientific quantities---such as transmission rates or ecological carrying capacities---by training on synthetic datasets where ground truth is known but inaccessible in real-world data.

\item \textbf{Cross-task generalization.} The SGNN approach can be applied across forecasting, regression, classification, and inference tasks.

\item \textbf{Mechanistic interpretability.} Through back-to-simulation attribution, SGNNs will identify which simulated regimes they internally associate with real-world inputs, offering insight into the underlying dynamics the model believes are active, and enabling process-level explanations.

\end{itemize}

\section{Applications and Results}

\subsection{Overview of Evaluation Strategy}

We evaluate Simulation-Grounded Neural Networks (SGNNs) across forecasting, inference, regression, and classification tasks spanning multiple scientific domains. Our experiments are designed to test the key capabilities of the SGNN framework.

\subsection{SGNNs Enable Robust Forecasting Across Domains}

\subsubsection{SGNNs Outperform CDC COVID-19 Forecasting Models}

We first evaluated SGNNs on real-world COVID-19 mortality forecasting during the early pandemic, using weekly death counts from all 50 U.S. states, Washington D.C., and Puerto Rico \cite{cdccovid}. This setting poses major challenges: limited early data, rapidly shifting transmission dynamics, evolving interventions, and substantial reporting noise.

The SGNN was pretrained entirely on large-scale mechanistic simulations, which incorporated a wide range of epidemic dynamics, including stochastic super-spreading, policy interventions, seasonality, underreporting, and surveillance delays (Appendix~\ref{appendix:simulations}). No real-world or COVID-specific data were used in pretraining.

We assessed performance using forecasting skill \cite{murphy1988skill, mittermaier2008impact, gneiting2007strictly}, defined as the percentage improvement in mean absolute error (MAE) relative to a naive baseline that predicts the most recent value:
\begin{equation}
\text{Skill} = 100 \times \left(1 - \frac{\text{MAE}_{\text{model}}}{\text{MAE}_{\text{baseline}}}\right)
\end{equation}
where the baseline prediction is $\hat{y}_{t+h} = y_t$ for all forecast horizons $h$. Normalizing by a baseline provides a dimensionless measure of skill, allowing us to compare and aggregate model performance across different locations and scales where absolute error magnitudes vary significantly. This is a stringent metric in infectious disease forecasting. The mean skill across all models in the CDC Forecast Hub was negative (--1.0) \cite{evaluationpnas}. For comparison, we benchmarked the SGNN against both the median skill (13.0) and best model's skill (34.0) of Forecast Hub models. We evaluate a full range of typical performance metrics for infectious disease performance, including weighted interval score and coverage, in appendix \ref{appendix:extended_data}.

\begin{figure}[ht!]
    \centering
    \includegraphics[width=\textwidth]{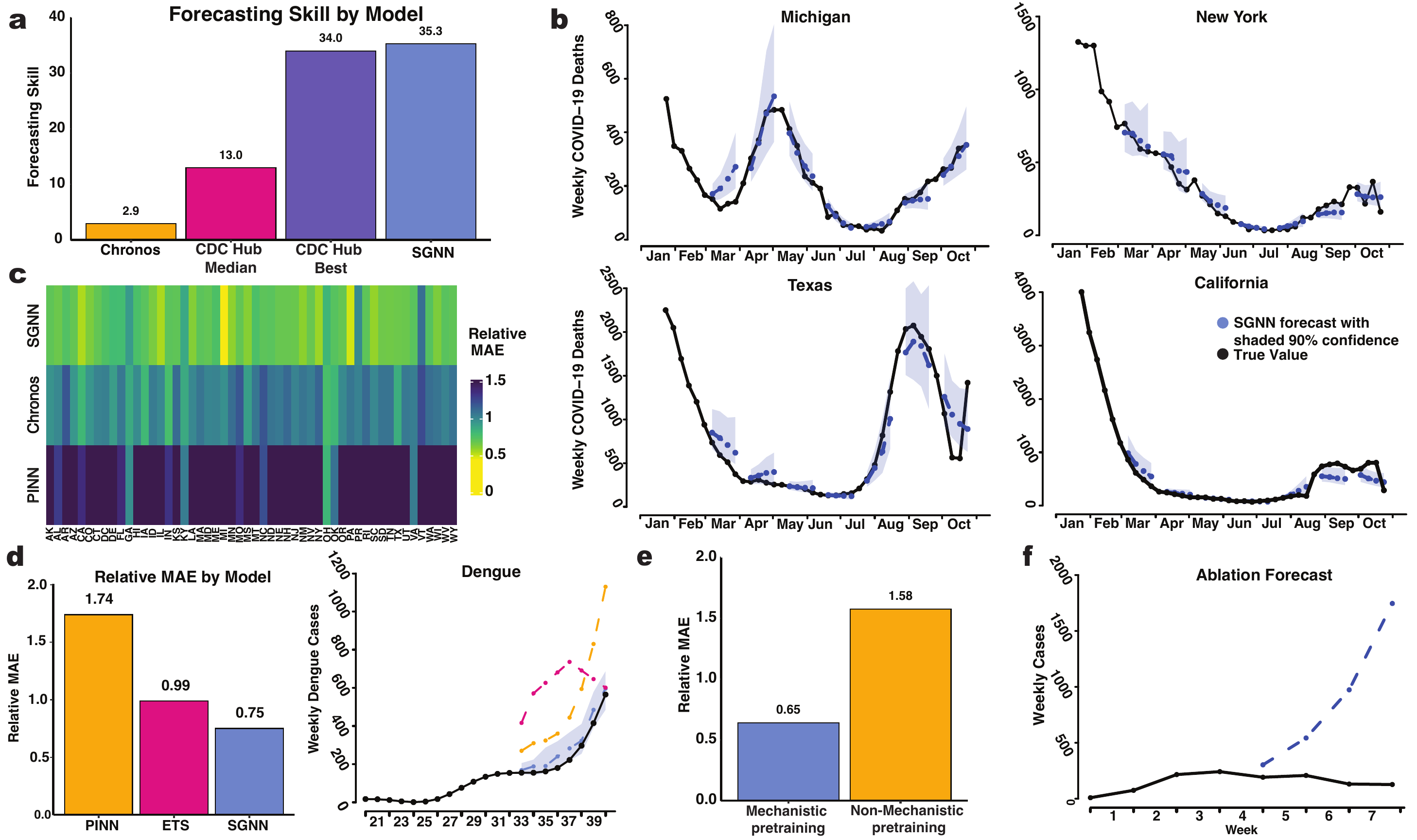}
    \caption{\textbf{SGNNs outperform state-of-the-art baselines in real-world disease forecasting and reveal the importance of mechanistic grounding.}
\textbf{(A)} SGNNs achieve 35.3\% forecasting skill on early COVID-19 mortality, almost tripling the CDC Forecast Hub median and exceeding its best model—despite using no real COVID-19 data.
\textbf{(B)} SGNNs produce accurate forecasts with calibrated uncertainty across diverse real-world locations, including Michigan, New York, Texas, and California (four states chosen to represent US regions: West Coast, East Coast, South, and Midwest).
\textbf{(C)} Across all U.S. states, SGNNs consistently achieve lower error (lighter) than Chronos and PINNs, showing robust generalization from synthetic pretraining.
\textbf{(D)} On dengue, a domain with fundamentally different transmission dynamics, SGNNs outperform both PINNs and statistical models in a fully zero-shot setting—demonstrating robustness to mechanistic misspecification.
\textbf{(E)} Replacing mechanistic simulations with neural simulators causes performance collapse, highlighting the necessity of mechanistic fidelity.
\textbf{(F)} Restricting pretraining to simple SEIR models leads to overconfident exponential forecasts, confirming that mechanistic diversity and surveillance realism are essential for robust downstream performance.}
    \label{fig:sgnn_disease_results}
\end{figure}

The SGNN outperformed the baselines (Figure~\ref{fig:sgnn_disease_results}). Compared to Chronos \cite{chronos}, a state-of-the-art foundation model pretrained on generic time series, the SGNN achieved a roughly 12-fold improvement in skill, underscoring the importance of mechanistic grounding. The SGNN also nearly tripled the Forecast Hub's median skill and slightly exceeded its best model performance. Finally, the SGNN outperformed a physics-informed neural network (PINN) baseline based on SEIR equations (skill = –112), which performed poorly in our tests in the presence of noise and data sparsity, consistent with prior observations \cite{rodriguez2022einns}. This demonstrates that incorporating domain knowledge into the data rather than constraining the output of a neural network to match specific equations may sometimes be a better strategy in the presence of large uncertainty and minimal real-world data.

\subsubsection{SGNNs Forecast Accurately Despite Mechanistic Misspecification}

Mechanistic models, whether traditional compartmental models or hybrid approaches like PINNs, rely on structural assumptions that may not hold in practice. Real-world systems often deviate from these assumptions, leading to brittle forecasts when the model form is even slightly wrong. This challenge is especially acute when key mechanisms (e.g., transmission routes or observation error) are misrepresented or omitted. To assess SGNNs under such conditions, we tested their forecasting performance with different mechanisms from those seen during pretraining.

We evaluated forecasting performance on real-world dengue outbreaks in Brazil \cite{tychodengue}. Dengue transmission involves mosquito vectors and mechanisms that are distinct from the human-to-human transmission used in SGNN pretraining simulations. Importantly, SGNNs were evaluated in a fully zero-shot setting: the model had no exposure to dengue data or vector-borne transmission processes during training.

To benchmark performance, we compared SGNNs against two baselines. The first was a PINN-style model trained directly on the dengue data using standard SEIR equations, imposing a rigid, incorrect mechanistic prior \cite{rodriguez2022einns}. The second was a purely statistical model based on exponential smoothing (ETS), which had access to the dengue data but incorporated no mechanistic assumptions.

SGNNs reduced forecasting error by 57\% relative to the PINN, demonstrating robustness under structural misspecification. SGNNs also outperformed the ETS model by 24\%, despite operating entirely zero-shot.

These findings underscore a strength of SGNNs: by training on a broad ensemble of mechanistic simulations spanning multiple model structures and parameter regimes, SGNNs learn flexible, scientifically grounded representations that can generalize when structural elements differ from those seen in training. This result suggests that SGNNs may not merely memorize surface patterns, but instead internalize scientific principles that extend across mechanistic boundaries. SGNNs may be learning abstract representations that capture core features of disease spread---such as delays, feedback, and depletion---that hold across different mechanistic forms.

\subsubsection{SGNNs Forecast Multispecies Ecological Dynamics Where Neural Nets Fail}

We evaluated SGNNs on ecological forecasting using two real-world datasets: (i) annual lynx and hare population data from the Hudson’s Bay Company (1821–1934), a low-dimensional predator-prey system \cite{R_datasets_lynx}, and (ii) multivariate butterfly species abundances from the UK Butterfly Monitoring Scheme (UKBMS), which tracks annual fluctuations for dozens of species \cite{UKBMS_2023}, posing a high-dimensional forecasting challenge.

Our SGNNs were pretrained on ecological simulations representing biological mechanisms and observational noise. For lynx-hare dynamics, training data were generated using stochastic Rosenzweig-MacArthur predator-prey models with demographic noise \cite{rosenzweig1963graphical}. For butterfly forecasting, simulations included logistic growth, weak interspecific competition, shared environmental noise, and observation processes such as negative binomial sampling. See Appendix~\ref{appendix:simulations} for full simulator details.

\begin{figure}
\centering
\includegraphics[width=0.625\linewidth]{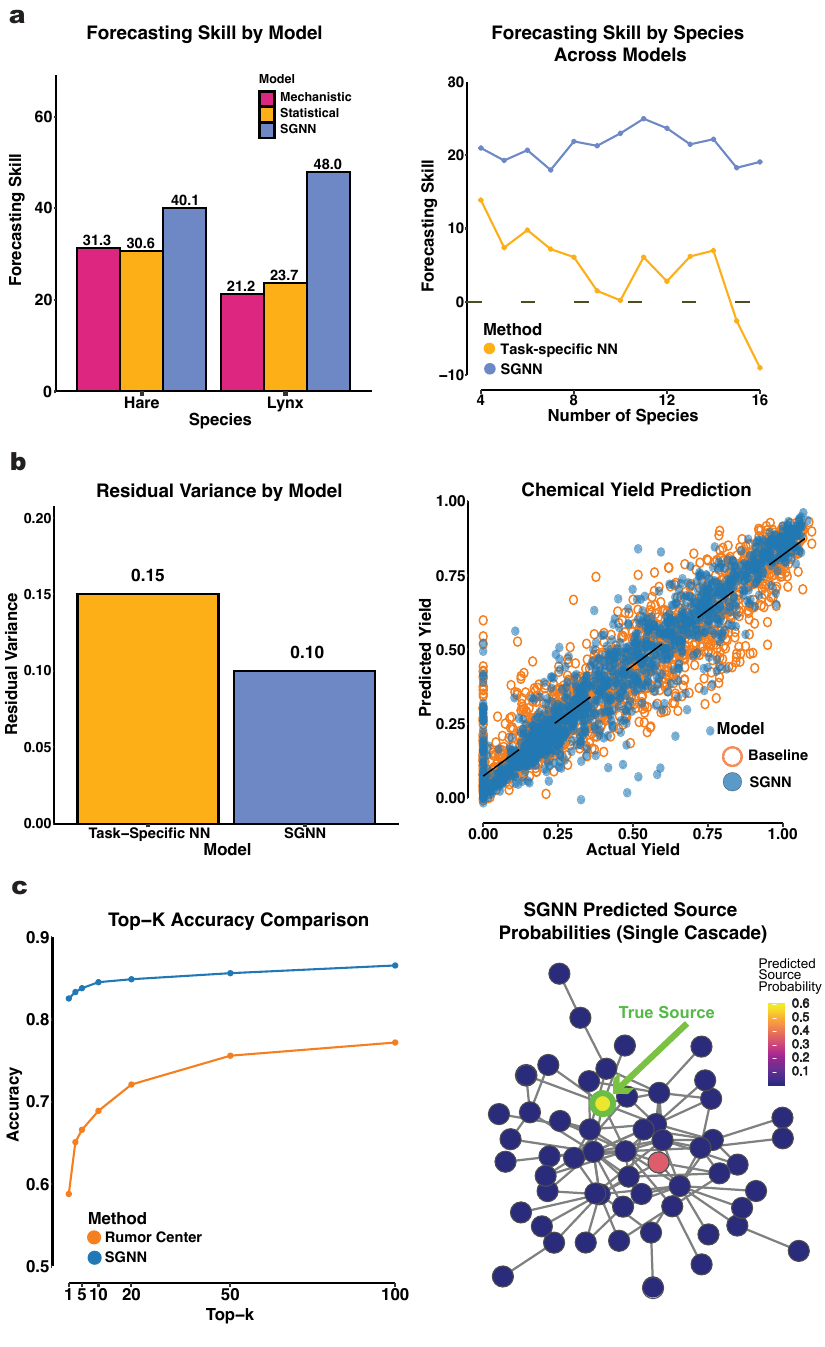}
\caption{\textbf{SGNNs generalize across scientific domains and task types.}
\textbf{(A)} \textit{Ecological forecasting:} SGNNs outperform task-specific neural networks on both low-dimensional predator-prey systems (hare and lynx) and high-dimensional multispecies forecasting from the UK Butterfly Monitoring Scheme. SGNNs maintain forecasting skill as the number of species increases, while baselines degrade sharply.
\textbf{(B)} \textit{Chemical yield prediction:} SGNNs reduce residual variance by one-third compared to state-of-the-art models trained directly on the Suzuki-Miyaura reaction dataset. Right: predicted vs.\ actual yield for SGNN (blue) and baseline (orange) models, showing tighter clustering around the identity line for SGNN.
\textbf{(C)} \textit{Diffusion source identification:} SGNNs accurately infer the source of information spread in partially observed cascades. Left: top-$k$ accuracy comparison shows SGNNs outperform the Rumor Center method. Right: predicted source probabilities for a representative cascade, with node color and size indicating model-assigned likelihood. SGNN correctly assigns high probability to the true source despite missing observations.
}
\label{fig:general_results}
\end{figure}

SGNNs achieved strong forecasting skill on both lynx and hare populations. As shown in Figure~\ref{fig:general_results}A (left), SGNNs outperformed statistical and mechanistic baselines, with forecasting skill of 48.0 for lynx and 40.2 for hare, higher than the 20.1-32.4 range for all baselines. These results highlight SGNNs ability to generalize to cyclic ecological systems from limited data.

On higher-dimensional forecasting, SGNNs maintained robust forecasting skill as the number of butterfly species increased from 4 to 16 (Figure~\ref{fig:general_results}A, right), outperforming task-specific neural networks (a neural network with the same architecture as the SGNN, but only trained on the historical data), which failed entirely in high dimensions. By 16 species, SGNN retained positive forecasting skill above 19.0, while baseline neural nets fell below zero. This illustrates how SGNNs can enable high-dimensional forecasting where standard approaches break down.

\subsection{Realistic, Diverse Mechanistic Grounding is Essential for Generalization}

To assess the importance of mechanistic grounding in SGNN training, we performed two ablation studies that isolate the contributions of both model realism and mechanistic diversity (Figure~\ref{fig:sgnn_disease_results}).

First, we replaced our simulation pipeline with a neural network trained directly on COVID-19 case data. This neural model acted as a simulator: it generated synthetic outbreaks by recursively forecasting forward from real historical contexts. We then trained a model with the same architecture as our SGNN directly on this simulated data. Despite the model having access to high amounts of simulated data, it was significantly outperformed by the SGNN trained on mechanistic simulations, yielding over twice the error (Figure \ref{fig:sgnn_disease_results}e). This experiment demonstrates that increasing data alone is insufficient. Mechanistic grounding is critical for stable and generalizable learning.

Second, we restricted the simulation corpus to classical SEIR models, removing all other mechanisms (e.g., asymptomatic spread, seasonality, super-spreading, stochasticity) and observational noise (e.g., underreporting, delays). These simulations assumed fixed parameters and homogeneous mixing, and omitted interventions or surveillance artifacts---factors critical for capturing real-world disease dynamics. This ablation isolates the impact of simplification: the underlying mechanistic structure remains broadly correct, but key features of realistic outbreaks are missing. As a result, SGNNs pretrained solely on SEIR data failed to learn these dynamics and produced systematically biased forecasts. For instance, during early COVID-19, they projected unmitigated exponential growth in locations where lockdowns had substantially curbed transmission (Figure \ref{fig:sgnn_disease_results}f). Performance dropped significantly relative to full SGNN pretraining, underscoring that both mechanistic diversity and observational realism are necessary for robust downstream generalization.

\subsection{SGNNs Boost Chemical Yield Regression with Minimal Pretraining}

We evaluated SGNNs on the nanomole-scale Suzuki-Miyaura reaction \cite{suzukimiyaura} dataset developed by Perera et al.~\cite{perera2018platform}, comprising 5,760 reactions spanning diverse combinations of ligands, bases, solvents, and aryl halides under automated flow chemistry. The task is challenging due to high-order interactions, failure-prone edge cases, and wide heterogeneity in reaction yields.

State-of-the-art data-driven models trained directly on this dataset leave 15\% of the variance unexplained on held-out test data \cite{granda2018controlling}. SGNNs, pretrained for only a few minutes on high-fidelity chemical reaction network simulations generated via a hybrid simulation pipeline (Appendix~\ref{appendix:simulations}), reduced residual variance to 10\%, a one-third reduction. This improvement reflects SGNN's ability to internalize mechanistic dependencies such as catalyst-substrate incompatibilities, reaction failures, and heteroscedastic noise that purely empirical models fail to capture from limited data alone. A major weakness of purely data-driven models was their inability to distinguish failed reactions, which often led to overconfident but inaccurate predictions in low-yield regimes (Figure~\ref{fig:general_results}B, right). By explicitly encoding failure probabilities into our synthetic simulator, SGNNs learned to recognize reactions likely to fail, leading to sharper, more calibrated predictions across the full yield distribution.
As shown in Figure~\ref{fig:general_results}B, SGNNs yield more accurate predictions with lower residual variance and better alignment with ground-truth yields.

\subsection{Supervised Inference for Unobservable Targets}

\subsubsection{SGNNs Accurately Infer the Source of Diffusion Cascades}

We evaluated SGNNs ability to identify the origin of diffusion in networks, a central challenge in computational social science, epidemiology, and security analytics. Given a noisy, partially observed cascade, the model must infer which node initiated the spread.

SGNNs were trained on synthetic cascades generated from the Independent Cascade model on Barabási–Albert networks \cite{barabasialbert}, with 20\% of infected nodes randomly masked to simulate incomplete observation. Despite this noise, SGNNs achieved high accuracy: 82.6\% top-1 accuracy and over 84.9\% top-20 accuracy, versus 58.8\% and 72.1\% for the Rumor Center method \cite{rumorcenter}. As shown in Figure~\ref{fig:general_results}C, SGNNs substantially outperform baselines on top-$k$ accuracy and produce spatially coherent, well-calibrated predictions across the network, even under missing data.

The SGNN does not approach 100\% top-1 accuracy because the true source is masked 20\% of the time, meaning that in one out of every five examples, the correct answer is effectively hidden. Yet the SGNN still achieves over 80\% accuracy, implying that it sometimes identifies the true source even when that node was never observed to be impacted. While some advanced inference methods can handle unobserved sources under specific assumptions \cite{zhai2015cascade}, SGNNs achieve this performance in a general setting without explicit source modeling, demonstrating an advance over heuristic or equation-based approaches.

\subsubsection{Estimating the Transmissibility of COVID-19}

Estimating the basic reproduction number ($R_0$)---the expected number of secondary infections generated by a single infectious individual in a fully susceptible population---is central to epidemic response and modeling \cite{anderson1992infectious, diekmann1990definition, fraser2007estimating}. However, $R_0$ is inherently unobservable, and traditional estimation methods rely on simplified assumptions often violated in practice, especially under data quality issues such as underreporting, delays, and early outbreak noise, as seen in early COVID-19.

To evaluate SGNNs, we pretrained on a diverse set of simulated outbreaks across multiple epidemiological models, with ground-truth $R_0$ values computed analytically \cite{vandenDriessche2002}. These simulations incorporated realistic noise and observation artifacts to mimic surveillance conditions (Appendix~\ref{appendix:simulations}).

\begin{figure}
\centering
\includegraphics[width=0.55\textwidth]{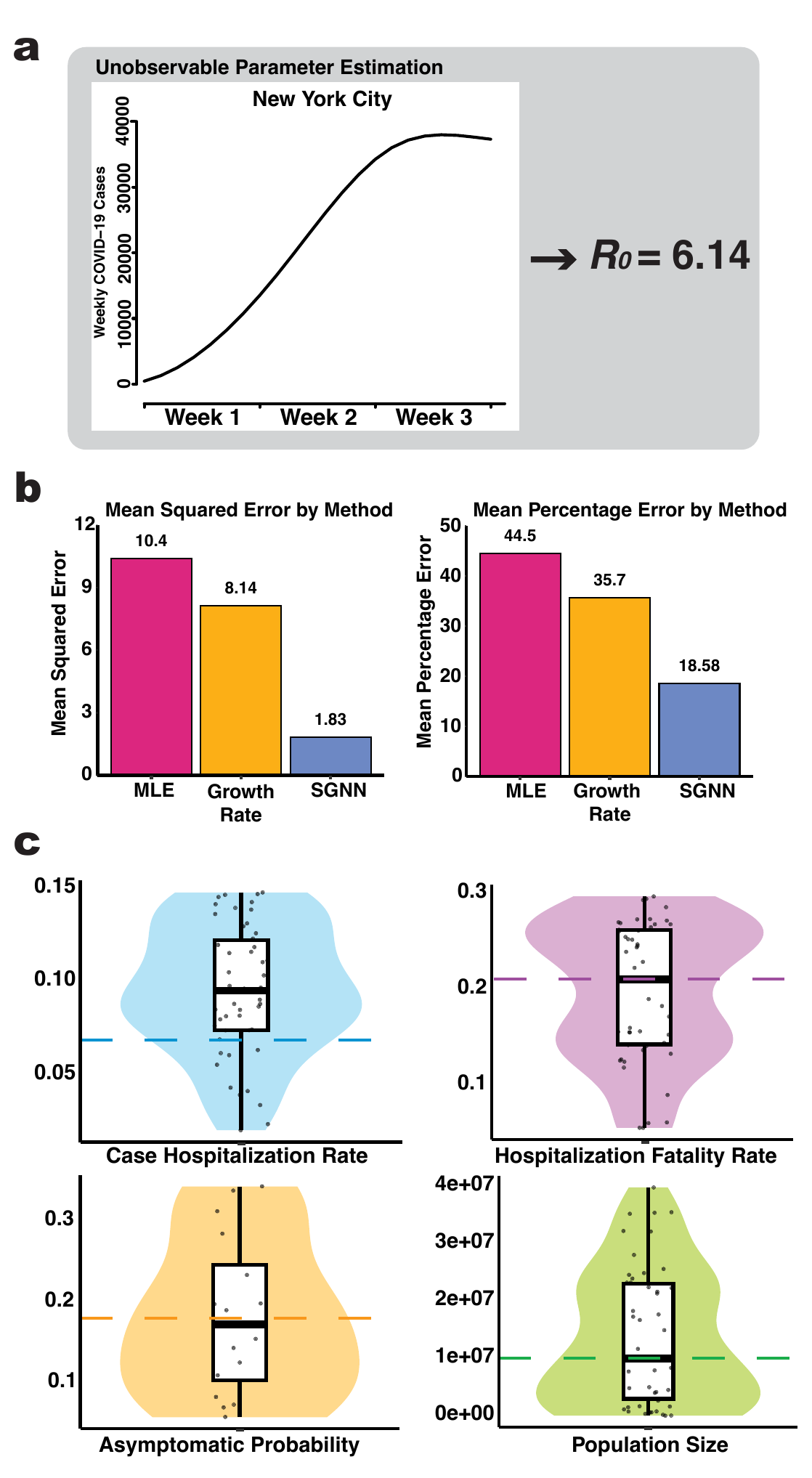}
\caption{\textbf{SGNNs accurately infer unobservable parameters and provide mechanistic interpretability via back-to-simulation attribution.}
\textbf{(A)} SGNNs infer a high reproduction number ($R_0 = 6.14$) for New York City from early COVID-19 case data (Feb–Mar 2020, shown in the plot), aligning with estimates from more complete datasets \cite{ives2020state}.
\textbf{(B)} SGNNs achieve substantially lower mean squared error (left) and mean percentage error (right) in $R_0$ estimation compared to maximum likelihood estimation (MLE) and exponential growth-based methods. Ground truth is known in this case since we are testing on simulated data.
\textbf{(C)} Back-to-simulation attribution retrieves the 50 most similar synthetic outbreaks to a real-world input and visualizes the distribution of their underlying mechanistic parameters. For Michigan COVID-19 mortality, the retrieved simulations align closely with true estimates for hospitalization rate, population size, asymptomatic transmission rate, and hospitalization fatality rate (dashed lines) \cite{griffin2024estimates}, confirming that the SGNN’s internal representations encode meaningful mechanistic structure.}
\label{fig:r0_results}
\end{figure}

We first tested the SGNN on synthetic outbreaks, comparing it to two baselines: (i) SIR-based maximum likelihood estimation (MLE), and (ii) exponential growth-based methods \cite{wallinga2007generation}. SGNN achieved an MSE of 1.83 and MPE of 18.6\%, outperforming MLE (10.4, 44.5\%) and growth-based estimates (8.1, 35.7\%) (Figure \ref{fig:r0_results}), demonstrating robustness to noise and model misspecification.

We next applied the SGNN to early real-world COVID-19 data (Feb 20-Mar 21, 2020) from U.S. states and cities. During this period, empirical estimates of $R_0$ were highly uncertain and often underestimated due to underreporting, inconsistent testing, and data delays. Many early methods assumed exponential growth and homogeneous mixing, neglecting asymptomatic spread and reporting lags.

The SGNN, trained solely on synthetic data, produced $R_0$ estimates from early COVID-19 data that were more consistent with later estimates once reporting had improved. For example, early estimates for New York City placed $R_0$ from 1.0-2.5 \cite{earlynyc}, yet subsequent studies using more complete data revised the estimate upward to over 5 \cite{vissat2022comparison, ives2020state}. As shown in figure \ref{fig:r0_results}, the SGNN, using only data from the early February-March period, correctly produced a high $R_0$ estimate for NYC (6.14), consistent with later, more complete analyses, versus 2.35 for exponential growth fitting. A full table of $R_0$ estimates is provided in Appendix \ref{appendix:extended_data}.

These results demonstrate that simulation-grounded training enables supervised learning for scientific quantities that are unobservable in real data and highly sensitive to noise and modeling assumptions. While only one demonstration, the SGNN’s robustness to data imperfections and its ability to implicitly infer complex transmission dynamics from short, noisy time series suggest it is well-suited for real-time epidemic inference.

\mbox{}

\subsection{Back-to-Simulation Attribution Enables Mechanistic Interpretability}
Because SGNNs are based in mechanistic models, they enable a new approach to mechanistic interpretability, back-to-simulation attribution, wherein we identify which simulated regimes the model deems most similar to a given real-world input. Let $f_\phi: \mathcal{X} \rightarrow \mathbb{R}^d$ denote the trained SGNN encoder that maps input time series to a $d$-dimensional embedding space. For a real-world observation $x$, we retrieve simulations $\{x_i\}$ from the training set $\mathcal{D}_{\text{syn}}$ by computing cosine similarity between $f_\phi(x)$ and $f_\phi(x_i)$, selecting the top-$k$ most similar embeddings.

This approach differs from direct time series comparison in two ways. First, retrieval in the fixed $d$-dimensional embedding space avoids the curse of dimensionality that would arise from comparing raw high-dimensional time series or parameter spaces. The embedding dimension remains constant regardless of forecast horizon or model complexity. Second, the retrieved simulations reflect the model's learned similarity rather than time series pattern matching. The encoder $f_\phi$ has been trained to map trajectories to representations that are predictive of outcomes. Thus, simulations with similar embeddings are likely the ones the model treats as mechanistically similar for the prediction task.

To perform this attribution, we passed the Michigan COVID-19 mortality time series through the trained SGNN encoder and extracted its embedding. We then computed cosine similarity between this embedding and all simulation embeddings in the synthetic training set. The top-50 most similar simulations were then analyzed to examine the distribution of their underlying mechanistic parameters.

Figure~\ref{fig:r0_results} shows violin plots of four key parameters across the top-50 retrieved simulations: case-hospitalization rate, hospitalization fatality rate, asymptomatic probability, and population size. These retrieved values align closely with known or plausible values for COVID-19 in Michigan, capturing the approximate population size (median population 10.1m, versus 10.14 actual), hospitalization and death probabilities consistent with empirical estimates, and realistic asymptomatic proportions \cite{griffin2024estimates}. This alignment confirms that the SGNN grounds its predictions in simulations that resemble real-world system dynamics.

Back-to-simulation attribution enables a different form of interpretability than feature attribution and parameter inference methods. Unlike SHAP or LIME \cite{shap, lime}, which identify which input features influenced a prediction, attribution reveals which mechanistic processes the model believes generated the data. Unlike simulation-based inference approaches that aim to provide statistically rigorous parameter estimation, back-to-simulation attribution explains the model's predictions by analyzing which mechanistic scenarios it considers similar to the observed data. This provides a domain-aligned window into the model's predictive reasoning while maintaining computational efficiency.

\section{Discussion}

By generating synthetic datasets from mechanistic simulators, the SGNN framework enables neural networks to learn both system dynamics and the observational processes that shape real-world data, including noise, underreporting, and delays.
Central to this framework is the shift from functional constraints to structural priors. While existing hybrid modeling approaches (e.g., PINNs or UDEs \cite{pinns, udes}) enforce domain knowledge by explicitly constraining a model’s derivative or loss function, simulation-grounded learning embeds scientific structure through examples. This allows the neural network to internalize the global manifold of physically plausible dynamics without being tied to a single, potentially incorrect mechanistic model structure. Because simulations span diverse model structures and parameter regimes, SGNNs develop a form of ensemble robustness, generalizing effectively even when real-world dynamics deviate from any single training simulator.

Training on simulated data also provides ground-truth labels for quantities that are unobservable in real settings, such as transmission rates or carrying capacities that lack direct measurements. This enables the integration of Simulation-Based Inference (SBI) \cite{sbi1, sbi2, sbi3}, which traditionally uses supervised learning for parameter inference, directly into the SGNN framework. While SBI often focuses on full Bayesian posterior estimation, our results demonstrate that SGNNs support a range of inference strategies. For example, our $R_0$ estimation experiments used direct quantile regression on synthetic outbreak trajectories with known ground-truth reproduction numbers, bypassing the need for explicit likelihood evaluation or posterior sampling while still producing calibrated uncertainty intervals. Similarly, the back-to-simulation attribution approach implicitly recovers distributional information over mechanistic parameters by retrieving the nearest synthetic neighbors in embedding space, providing a nonparametric alternative to formal posterior inference. Depending on the scientific need, one can move from full posterior estimation to these computationally efficient alternatives, all while maintaining the scientific grounding provided by the synthetic training corpus. In this sense, SGNNs serve as a general vehicle for SBI, allowing these techniques to be applied to complex forecasting and regression tasks where they were previously isolated.

Beyond parameter inference, back-to-simulation attribution also provides a distinct form of mechanistic interpretability. Unlike post-hoc feature attribution methods that identify which input variables influenced a prediction, the retrieved simulations reveal which mechanistic regimes the model associates with observed data. For example, if a model retrieves simulations characterized by high asymptomatic transmission when shown an outbreak with few reported cases, it suggests the model has internally identified a regime of substantial hidden spread. This provides actionable, process-level insight, allowing practitioners to validate the model's reasoning against domain knowledge and use those mechanistic insights to guide intervention strategies.

Our results suggest that simulation-grounded learning can capture generalizable representations of scientific structure that extend beyond the training distribution. The success of this paradigm has since been further validated through the development of Mantis \cite{mantis}, a more advanced simulation-grounded foundation model for infectious disease forecasting. By scaling the SGNN framework to a massive synthetic corpus, Mantis achieves performance competitive with state-of-the-art models across diseases and geographies entirely out-of-the-box, without requiring real-world fine-tuning.

Despite these strengths, SGNNs have important limitations that we plan to explore further in future work. First, performance depends fundamentally on simulation quality: if key mechanisms are absent from the training simulators, performance can be substantially degraded. While training on diverse simulator ensembles provides robustness to individual model misspecification, systematic gaps across all training simulations will propagate to predictions, underscoring the importance of careful choices around the mechanistic models and parameter ranges used for training. Even with realistic noise models and observational artifacts, we anticipate that systematic differences between synthetic and real data will hurt performance. Also, we lack theoretical guarantees about when SGNNs will generalize. While our results show promising zero-shot transfer (e.g., human-to-human models predicting vector-borne disease), we cannot predict \textit{a priori} which cross-domain generalizations will succeed. Second, generating comprehensive simulation corpora can be computationally expensive, particularly for high-fidelity mechanistic models. Though this is a one-time cost and inference remains fast, it may be prohibitive for extremely expensive simulators. Finally, back-to-simulation attribution reveals what the model believes is happening mechanistically, but that can be wrong. Retrieved simulations reflect learned associations that, while often informative, can be incorrect and depend heavily on the simulation prior. Understanding when simulation-grounded learning succeeds and when it fails remains an important direction for future work.

\section{Author Contributions}
CD led the study and developed the code and initial draft of the manuscript. CH assisted with initial code development and RM developed figure visualizations. MCE, CD, CH, and RM all contributed substantial edits and revisions for the final manuscript. 

\bibliographystyle{unsrt}
\bibliography{references}

\begin{appendices}

\section{Simulation Frameworks}\label{appendix:simulations}

\subsection{Infectious Disease Simulation Framework}

\paragraph{Mechanistic simulator.} To generate synthetic training data for epidemiological forecasting and inference tasks, we developed a universal infectious disease simulator that synthesizes diverse outbreak trajectories across a wide range of plausible regimes. This simulator flexibly instantiates classical compartmental models—such as Susceptible-Infectious-Recovered (SIR), Susceptible-Exposed-Infectious-Recovered (SEIR), and Susceptible-Exposed-Asymptomatic-Infectious-Recovered (SEAIR)—augmented with stochasticity, seasonal variation, demographic turnover, and realistic reporting artifacts. The result is a rich corpus of simulated epidemics that capture both mechanistic diversity and observational realism, serving as the foundation for SGNN pretraining.

Each simulation models a well-mixed population of size $N \sim \text{LogUniform}(5 \times 10^4, 5 \times 10^7)$. The compartmental structure is chosen randomly: 70\% of simulations include an exposed compartment ($E$), and 50\% include an asymptomatic compartment ($A$), resulting in SIR, SEIR, or SEAIR dynamics. The full SEAIR model is governed by:

\[
\begin{aligned}
\frac{dS}{dt} &= -\lambda(t) S + \omega R + \mu N - \mu S, \\
\frac{dE}{dt} &= \lambda(t) S - \sigma E - \mu E, \\
\frac{dA}{dt} &= p_A \sigma E - \gamma A - \mu A, \\
\frac{dI}{dt} &= (1 - p_A) \sigma E - \gamma I - \mu I, \\
\frac{dR}{dt} &= \gamma I + \gamma A - \omega R - \mu R,
\end{aligned}
\]

where parameters are sampled per simulation as follows:

\vspace{0.5em}
\begin{tabular}{lcl}
\textbf{Parameter} & \textbf{Distribution} & \textbf{Description} \\
\hline
$\beta$ & $\sim \text{Uniform}(0.10, 1.00)$ & Base transmission rate per contact \\
$\gamma$ & $\sim \text{Uniform}(0.10, 0.33)$ & Recovery rate \\
$\sigma$ & $\sim \text{Uniform}(0.20, 0.40)$ & Latency rate (if exposed state present) \\
$\omega$ & $\sim \text{Uniform}(0.001, 0.0075)$ & Waning immunity rate (if enabled) \\
$\mu$ & $\sim \text{Uniform}(0, 1/365)$ & Birth/death rate (if enabled) \\
$p_A$ & $\sim \text{Uniform}(0.1, 0.7)$ & Fraction asymptomatic (if enabled) \\
$\alpha$ & $\sim \text{Uniform}(0.3, 1.0)$ & Relative transmissibility of $A$ vs. $I$ \\
\end{tabular}
\vspace{0.5em}

Transmission includes super-spreading and seasonal forcing. The time-varying force of infection is:

\[
\lambda(t) = \beta(t) \cdot \left( \frac{I(t) + \alpha A(t)}{N(t)} \right) \cdot s(t) \cdot \eta(t),
\]

where $s(t)$ is a sum of 1–4 sinusoidal harmonics (with random amplitudes $\sim \text{Uniform}(0.05, 0.20)$ and phases), and $\eta(t)$ is gamma-distributed individual-level overdispersion to model super-spreading. Specifically, $\eta \sim \text{Gamma}(k, k)$ for $k \in [0.1, 1.0]$ in a subset of simulations.

To simulate multi-wave epidemics, $\beta(t)$ may shift at 1–5 change points, each with different values sampled from the same $\beta$ distribution. Clinical severity (hospitalization and death probabilities) also varies across waves.

Stochasticity is implemented via discrete-time simulations with daily time steps. For each day, new transitions between compartments are sampled using binomial draws based on the current state and transition probabilities derived from the rates above. This approximates tau-leaping dynamics.

\paragraph{Interventions.} Non-pharmaceutical interventions (NPIs) are included in 25\% of simulations. NPIs are triggered when new case counts exceed a sampled threshold $\sim \text{Uniform}(0.001N, 0.01N)$. Interventions reduce $\beta(t)$ by a sampled factor $\sim \text{Uniform}(0.2, 0.6)$, lasting for $\sim \text{Uniform}(14, 120)$ days. They are relaxed when case counts fall below a separate threshold for several days. This simulates lockdowns or behavior change in response to outbreak dynamics.

\paragraph{Demographic turnover and importation.} In 80\% of simulations, birth and death processes are active, with Poisson-distributed inflow into $S$ and deaths from all compartments. Importations into $E$ or $I$ occur randomly with daily rate $\sim \text{LogUniform}(0.0001, 0.01)$, supporting both isolated and repeatedly seeded outbreaks. If waning immunity is enabled, recovered individuals return to $S$ at rate $\omega$, supporting endemic cycling.

\paragraph{Clinical outcomes.} Hospitalizations and deaths are sampled from symptomatic infections using per-wave probabilities:
\begin{itemize}
  \item Hospitalization: $\sim \text{Uniform}(0.02, 0.15)$
  \item Death given hospitalization: $\sim \text{Uniform}(0.05, 0.30)$
\end{itemize}
Delays from infection to hospitalization and to death are sampled from gamma distributions with:
- Hospitalization delay: mean $\sim \text{Uniform}(5, 12)$ days
- Death delay: mean $\sim \text{Uniform}(14, 21)$ days

\paragraph{Observation model.} Realistic surveillance effects are applied to the ground-truth time series. These include:
\begin{itemize}
  \item \textbf{Reporting delay}: geometric delay kernels with mode $\sim \text{Uniform}(0, 3)$ days
  \item \textbf{Underreporting}: case detection probability increases logistically from $\sim \text{Uniform}(5\%, 40\%)$ to $\sim \text{Uniform}(25\%, 85\%)$ over time, with logistic midpoints drawn uniformly over the simulation window
  \item \textbf{Weekday effects}: multiplicative effects $\sim \text{Normal}(1.0, 0.05)$ applied to each weekday
  \item \textbf{Noise}: lognormal multiplicative noise applied to cases (high), hospitalizations (medium), and deaths (low)
\end{itemize}

\paragraph{Outputs.} Each simulation returns:
\begin{itemize}
  \item True and reported daily time series for infections, hospitalizations, and deaths
  \item True latent trajectories ($S, E, I, A, R$), daily $R_t$, and clinical outcomes
  \item Intervention metadata (timing, effect size), simulation parameters, and noise specifications
\end{itemize}

This end-to-end framework yields a high-diversity corpus of simulated epidemics with realistic mechanistic and observational properties, enabling generalizable pretraining of SGNNs across infectious disease tasks.

\begin{table}[H]
\centering
\small
\begin{tabular}{>{\raggedright}p{2.5cm} p{4cm} p{6.5cm}}
\toprule
\textbf{Parameter} & \textbf{Distribution / Range} & \textbf{Description} \\
\midrule
\multicolumn{3}{l}{\textbf{Transmission and Progression}} \\
\midrule
$\beta$ & $\text{Uniform}(0.10, 1.00)$ & Effective transmission rate per contact \\
$\gamma$ & $\text{Uniform}(0.10, 0.33)$ & Recovery rate from infection \\
$\sigma$ & $\text{Uniform}(0.20, 0.40)$ & Rate of progression from exposed to infectious \\
$p_A$ & $\text{Uniform}(0.10, 0.70)$ & Fraction of cases that are asymptomatic \\
$\alpha$ & $\text{Uniform}(0.30, 1.00)$ & Relative transmissibility of asymptomatic individuals \\
$k$ (dispersion) & $\text{Uniform}(0.10, 1.00)$ (in some sims) & Overdispersion for super-spreading (used in $\eta \sim \text{Gamma}(k,k)$) \\
$s(t)$ & 1–4 harmonics, amp $\sim \text{U}(0.05, 0.20)$ & Seasonal forcing using sinusoidal harmonics \\
\midrule
\multicolumn{3}{l}{\textbf{Demography and Immunity}} \\
\midrule
$N$ & $\text{LogUniform}(5\times10^4, 5\times10^7)$ & Total population size \\
$\mu$ & $\text{Uniform}(0, 1/365)$ & Per capita birth and death rate (if enabled) \\
$\omega$ & $\text{Uniform}(0.001, 0.0075)$ & Waning immunity rate (R $\rightarrow$ S) \\
Importation rate & $\text{LogUniform}(0.0001, 0.01)$ & Daily chance of importation into $E$ or $I$ \\
\midrule
\multicolumn{3}{l}{\textbf{Clinical Outcomes}} \\
\midrule
Hosp. prob. & $\text{Uniform}(0.02, 0.15)$ & Probability of hospitalization (per wave) \\
Death | Hosp. & $\text{Uniform}(0.05, 0.30)$ & Probability of death conditional on hospitalization \\
Hosp. delay & $\text{Gamma}(\mu, \theta)$; $\mu \sim \text{Uniform}(5,12)$ & Delay from infection to hospitalization \\
Death delay & $\text{Gamma}(\mu, \theta)$; $\mu \sim \text{Uniform}(14,21)$ & Delay from infection to death \\
\midrule
\multicolumn{3}{l}{\textbf{Intervention (NPI) Parameters}} \\
\midrule
Trigger threshold & $\text{Uniform}(0.001N, 0.01N)$ & Cases required to trigger intervention \\
Relaxation threshold & $\text{Uniform}(0.0002N, 0.002N)$ & Cases required to relax intervention \\
Reduction factor & $\text{Uniform}(0.20, 0.60)$ & Reduction in $\beta(t)$ during intervention \\
Duration & $\text{Uniform}(14, 120)$ (days) & Minimum intervention duration \\
\midrule
\multicolumn{3}{l}{\textbf{Observation Model (Reporting)}} \\
\midrule
Initial report rate & $\text{Uniform}(0.05, 0.40)$ & Case reporting probability early in epidemic \\
Final report rate & $\text{Uniform}(0.25, 0.85)$ & Case reporting probability later in epidemic \\
Logistic midpoint & $\text{Uniform}(0.2, 0.7)$ of time axis & Center point for logistic increase in reporting \\
Delay kernel mode & $\text{Uniform}(0, 3)$ (days) & Mode of geometric distribution for reporting delay \\
Weekday effects & $\text{Normal}(1.0, 0.05)$ & Multiplicative weekday effects (e.g., weekend drop) \\
Case noise & $\text{LogNormal}(0, \sigma^2), \sigma \sim \text{U}(0.15, 0.25)$ & Multiplicative noise applied to case counts \\
Hosp noise & $\text{LogNormal}(0, \sigma^2), \sigma \sim \text{U}(0.10, 0.15)$ & Noise applied to hospitalization counts \\
Death noise & $\text{LogNormal}(0, \sigma^2), \sigma \sim \text{U}(0.05, 0.10)$ & Noise applied to death counts \\
\bottomrule
\end{tabular}
\caption{Summary of stochastic simulation parameter distributions used to generate mechanistically diverse, observationally realistic synthetic outbreaks.}
\label{tab:sim-params}
\end{table}

\subsection{Ecological Simulation Frameworks}

\subsubsection{Butterfly Simulation Framework}

\paragraph{Theoretical model.} To generate synthetic multivariate population dynamics for the butterfly forecasting experiments, we developed a hybrid mechanistic simulation framework that blends ecological growth models, interspecific competition, environmental stochasticity, and realistic observation processes. This framework is designed to approximate both the underlying biological drivers of species abundance fluctuations and the statistical properties of survey-based monitoring data.

Each simulation begins by sampling a random number of species (up to 32), with species-specific growth parameters drawn independently. The intrinsic growth rate for species $i$, denoted $r_i$, is drawn from a uniform distribution $r_i \sim \mathcal{U}(0.15, 0.4)$. Baseline carrying capacities $K_i$ are assigned as scaled multiples of initial population sizes, with $K_i = N_{0i} \times U(1.5, 2.5)$, where $N_{0i} \sim 10^{\mathcal{U}(1.7, 2.4)}$. This yields baseline abundances ranging from approximately 50 to 250 individuals per species. Interspecific competition is modeled using a Lotka-Volterra framework \cite{lotka2002contribution}, with pairwise competition coefficients $\alpha_{ij}$ drawn from a truncated normal distribution $\alpha_{ij} \sim \mathcal{N}(0.03, 0.01)$, clipped to be non-negative. Self-competition terms $\alpha_{ii}$ are set to zero. This results in weak but widespread competition, consistent with the diffuse resource-mediated interactions observed in real butterfly communities. 

Environmental variability is introduced through two distinct mechanisms. First, longitudinal environmental shocks are modeled via an annual AR(1) process that generates temporally correlated environmental multipliers. The initial value is drawn from $E_0 \sim \mathcal{N}(1.0, 0.05)$, and subsequent years evolve as $E_t = 0.7 \cdot E_{t-1} + \epsilon_t$, where $\epsilon_t \sim \mathcal{N}(0, 0.05)$. This process is exponentiated and normalized to yield multiplicative factors applied to carrying capacities over time. Second, seasonal forcing is applied via sinusoidal modulation of carrying capacities to simulate annual cycles. Specifically, we define seasonally adjusted carrying capacity as:

\[
K_{it} = K_i \cdot E_t \cdot \left[1 + A_s \sin\left(2\pi \frac{t + \phi}{12}\right)\right],
\]

where the seasonal amplitude $A_s = 0.15$ and the phase offset $\phi$ is drawn from $\mathcal{U}(0, 1)$.

\paragraph{Real-world effects.} After integrating the deterministic dynamics, process noise is applied to simulate intrinsic demographic stochasticity. For each species and time point, we perturb the latent population size with Gaussian noise centered at zero and with standard deviation equal to 3\% of the population value. These noisy dynamics are generated by solving a system of ordinary differential equations that govern coupled species growth and competition. For species $i$, the governing equation is:

\[
\frac{dN_i}{dt} = r_i N_i \left(1 - \frac{N_i + \sum_{j} \alpha_{ij} N_j}{K_{it}} \right),
\]

where $K_{it}$ reflects the time-varying carrying capacity influenced by environmental and seasonal variation. The system is integrated numerically over a 100-year simulation horizon using a time step of one year via `scipy.integrate.odeint`.

To simulate realistic observational data, we apply a multi-stage noise process to the latent population trajectories. First, each true abundance $N_{it}$ is converted to an observed count using a negative binomial distribution with high dispersion: $y_{it} \sim \mathrm{NB}(\mu = N_{it}, \text{overdispersion} = 2000)$. This models sampling error and overdispersion typical in ecological field surveys. The resulting counts are then transformed to log base 10 and perturbed with additive Gaussian noise of standard deviation 0.08 to simulate additional measurement error, rounding artifacts, and variability introduced during reporting and digitization.

The final output consists of multivariate, log-transformed species abundance time series, representing realistic synthetic training data for ecological forecasting tasks. These data reflect complex population dynamics shaped by nonlinear growth, weak competition, temporally correlated environmental shocks, seasonal cycles, demographic noise, sampling error, and layered observational uncertainty.

\subsubsection{Lynx-Hare Simulation Framework}

To simulate predator-prey cycles reflective of historical lynx-hare dynamics, we implemented a stochastic extension of the Rosenzweig-MacArthur model \cite{rosenzweig1963graphical}, incorporating a mild self-limitation term on predator growth. The model consists of two coupled differential equations describing the dynamics of the hare ($H$) and lynx ($L$) populations:

\[
\frac{dH}{dt} = r H \left(1 - \frac{H}{K} \right) - \beta H L
\]
\[
\frac{dL}{dt} = \delta H L - \gamma L - \rho L^2,
\]

where $r$ is the intrinsic growth rate of hares, $K$ is the hare carrying capacity, $\beta$ is the predation rate, $\delta$ is the conversion efficiency from prey to predator, $\gamma$ is the natural mortality rate of lynx, and $\rho$ is a density-dependent predator self-limitation term. To ensure stable oscillatory dynamics, parameters are sampled from broad but biologically plausible ranges: $r \sim \mathcal{U}(0.4, 0.6)$, $K \sim \mathcal{U}(80, 120)$, $\beta \sim \mathcal{U}(0.02, 0.04)$, $\delta \sim \mathcal{U}(0.025, 0.04)$, $\gamma \sim \mathcal{U}(1.0, 2.0)$, and $\rho \sim \mathcal{U}(0.0005, 0.002)$.

Stochasticity is introduced during forward iteration by injecting Gaussian demographic noise into both populations at each time step. The noise magnitude is scaled inversely with population size, reflecting the empirical observation that larger populations exhibit lower relative variance. Specifically, the coefficients of variation are computed as:

\[
\mathrm{CV}_H = \min\left(0.1, \frac{200}{1000 \cdot H}\right), \quad \mathrm{CV}_L = \min\left(0.1, \frac{100}{1000 \cdot L}\right).
\]

Population updates are drawn from normal distributions with these variances, while ensuring all values remain non-negative and bounded by ecologically plausible maxima ($H_{\max} = 200$, $L_{\max} = 80$).

Final simulation outputs consist of annual time series for hare and lynx populations over a 100-year period. To match the structure of real-world empirical records such as fur trapping data, outputs are expressed in "pelt-unit" scaling and passed through the same noise pipeline as the butterfly model to ensure realistic training data for low-dimensional predator-prey forecasting tasks.

\subsection{Synthetic Data Generation for Suzuki-Miyaura Cross-Coupling}

To support pretraining in the domain of small-molecule synthesis, we constructed a hybrid simulation framework for the Suzuki–Miyaura cross-coupling reaction \cite{suzukimiyaura}—a palladium-catalyzed process widely used in pharmaceutical and materials chemistry to form carbon–carbon bonds between aryl halides and boronates~\cite{suzukimiyaura}. Our simulator was designed to produce synthetic reaction yields that match the structural patterns and failure behavior observed in the empirical dataset of Perera et al.~\cite{perera2018platform}, which includes 5,760 nanomole-scale reactions executed on an automated flow synthesis platform.

While our simulator does not explicitly solve chemical reaction networks via ODEs or stochastic kinetics, it is nonetheless grounded in mechanistic principles. This design reflects the nature of the available data: the experimental dataset reports only final reaction outcomes (i.e., isolated yields), not time-resolved concentration trajectories. As such, our focus is on modeling endpoint behavior rather than simulating reaction dynamics. We embed domain knowledge through structured component effects, chemically motivated interaction terms, and expert-informed failure rules that reflect known catalytic behaviors (e.g., ligand–substrate incompatibilities, base–boronate failures). The result is a scalable generative framework that captures core mechanistic dependencies without requiring first-principles simulation, enabling effective pretraining of SGNNs on high-fidelity synthetic data.

\subsection*{Simulation Design}

We modeled reaction yield as a structured function of five categorical inputs: \textit{aryl halide}, \textit{boronate}, \textit{ligand}, \textit{base}, and \textit{solvent}. For each component \( x \) in category \( C \), we estimated a main effect \( \mu_x \) as:
\begin{equation}
\mu_x = \mathbb{E}[\text{Yield} \mid x] - \mathbb{E}[\text{Yield}]
\end{equation}
These were combined to form a base prediction:
\begin{equation}
\hat{y}_{\text{base}} = \mu + \sum_{C} \mu_{x_C}
\end{equation}
where \( \mu \) is the global yield mean and \( x_C \) is the component selected from category \( C \).

To capture higher-order dependencies, we added:
\begin{itemize}
    \item \textbf{Pairwise interactions} \( \delta_{x_C, x_{C'}} \) for combinations with sufficient empirical support (e.g., aryl halide–ligand, base–boronate).
    \item \textbf{Three-way interactions} \( \delta_{x_A, x_B, x_C} \) for known mechanistic phases (e.g., oxidative addition: aryl halide–ligand–solvent).
    \item \textbf{Exact memorization} for all reaction tuples observed in the Perera dataset, storing their empirical mean and variance for maximal fidelity.
\end{itemize}

This yields a composite predictive model:
\begin{equation}
\hat{y} = \hat{y}_{\text{base}} + \sum \delta_{\text{pair}} + \sum \delta_{\text{3way}} + \epsilon
\end{equation}

\subsection*{Failure Modeling}

Reactions with yields below 5\% were treated as failures. We estimated the probability of failure using a hybrid method:
\begin{itemize}
    \item Empirical failure rates computed per component.
    \item Heuristics such as:
    \begin{itemize}
        \item Strong base + weak ligand (e.g., NaOtBu + PPh$_3$)
        \item BF$_3$K boronates with weak bases
        \item Aryl chlorides with low-activity ligands
    \end{itemize}
\end{itemize}

The final failure probability \( p_{\text{fail}} \) was computed as a weighted sum of these effects. Reactions sampled as failures were assigned yields drawn from a uniform distribution in the interval $[0.001, 0.04]$.

\subsection*{Heteroscedastic Noise and Calibration}

For successful reactions, we modeled residual uncertainty using a heteroscedastic Gaussian noise model:
\begin{equation}
y \sim \mathcal{N}(\hat{y}, \sigma^2(\hat{y}))
\end{equation}
The variance function \( \sigma^2(\hat{y}) \) was learned using residuals from the real dataset, stratified into 40 equal-count bins across predicted yield levels. This captures the increased uncertainty at intermediate-to-high yields and tighter calibration at the extremes. All sampled yields were then rescaled to match the empirical mean and standard deviation via z-score normalization, and clipped to the physical interval $[0.001, 0.999]$.

\subsection*{Sampling Strategy}

To ensure broad yet chemically meaningful coverage of the reaction space, we used a stratified sampling protocol:
\begin{itemize}
    \item 60\% of reactions drawn from exact empirical combinations (memorized in the dataset)
    \item 30\% from high-yielding partial combinations (e.g., known aryl halide–ligand pairs paired with new bases/solvents)
    \item 10\% uniformly sampled across the full combinatorial space
\end{itemize}

This strategy preserves empirical structure while encouraging generalization to underexplored regions.

\subsection*{Dataset Summary}

We generated a total of 2.5 million synthetic reactions. After calibration, the final dataset statistics were:

\begin{center}
\begin{tabular}{l|c}
\textbf{Metric} & \textbf{Value} \\
\hline
Unique synthetic reactions & 2,500,000 \\
Mean yield & 0.62 \\
Standard deviation & 0.28 \\
Failure rate (yield $<$ 5\%) & 9.1\%
\end{tabular}
\end{center}

This synthetic corpus preserves the structural complexity, sparsity, and statistical patterns of real-world reaction optimization, while enabling scalable simulation-grounded learning in low-data chemical settings.

\subsection{Synthetic Cascade Generation for Source Identification}

To train SGNNs for source identification, we generated synthetic diffusion cascades using the Independent Cascade (IC) model on a fixed Barabási–Albert graph \cite{barabasialbert} with 1,000 nodes, where each node adds 5 edges. The IC model captures a key property of real-world information diffusion: once activated, each node has a single chance to infect each of its neighbors with probability $p = 0.05$, proceeding in discrete time steps. Each cascade begins with a randomly selected source node and unfolds for up to 15 time steps or until the process terminates.

For each simulation, we recorded the time of infection for every node. To reflect realistic observational constraints—such as incomplete contact tracing or partial social media logs—we randomly masked 20\% of infected nodes by setting their infection time to a sentinel value, indicating that they were unobserved. Nodes not infected during the cascade were also assigned this sentinel value. The resulting input for each node consisted of its infection time (observed or unobserved), paired with structural encodings describing its location in the network (described in the model architecture).

This setup produces a large corpus of partially observed cascades with known ground-truth sources, enabling supervised training of SGNNs to learn latent causal structure under uncertainty.

\section{SGNN Architecture and Training Details}\label{appendix:architecture}

\subsection{Model Architecture}

To demonstrate the flexibility of the SGNN framework across tasks and domains, we implemented a range of neural architectures tailored to each problem type. We found that straightforward, task-appropriate models performed well out-of-the-box when trained with simulation-grounded supervision. This suggests that the SGNN training framework reduces reliance on choosing a specific architecture. We selected architectures that align naturally with each task structure: e.g., sequence models for time series and graph models for networks.

\paragraph{Time Series Forecasting.}
For epidemiological and ecological forecasting, we used two classes of models:

\begin{itemize}
\item \textbf{PatchTST-style temporal patching transformers} \cite{patchtst}: These models divide input sequences into non-overlapping temporal patches (e.g., 7-day blocks) and encode each as a token, allowing efficient self-attention over longer sequences. This design captures coarse-grained temporal structure while preserving positional resolution. 
\item \textbf{Hybrid convolutional neural network (CNN{)}-Transformer architectures}: We also employed models that combine convolutional feature extraction with transformer-based sequence modeling. CNN layers detect local patterns (e.g., sudden surges, weekly reporting artifacts), while transformers model long-range dependencies and dynamics. This hybrid design balances expressivity and efficiency, particularly for noisy, high-variance real-world time series with both short-term and long-term patterns.
\end{itemize}

Both architectures are capable of producing calibrated probabilistic forecasts when trained using quantile loss. Importantly, neither model uses domain-specific feature engineering or hand-crafted mechanisms—their strong performance is driven entirely by the structure learned from simulation-based pretraining.

\paragraph{Network Diffusion Source Identification.}
To infer the origin of a diffusion cascade from noisy, partially observed data, we implemented a Graph Attention Network (GAT) \cite{gat} enhanced with spectral positional encodings. This architecture allows the model to leverage both the temporal dynamics of the cascade and the topology of the underlying network.

\begin{itemize}
\item \textbf{Input Representation and Laplacian PE:} Each node $v$ is initialized with a feature vector $\mathbf{h}_v = [t_v || \mathbf{p}_v]$, where $t_v$ is the observed infection time (or a sentinel value for unobserved nodes) and $\mathbf{p}_v \in \mathbb{R}^{16}$ is a Laplacian Positional Encoding (LapPE) \cite{laplacianpe}. To compute $\mathbf{p}_v$, we define the symmetric normalized Laplacian $\mathcal{L} = \mathbf{I} - \mathbf{D}^{-1/2}\mathbf{A}\mathbf{D}^{-1/2}$. We perform an eigendecomposition of $\mathcal{L}$ and retain the first $k=16$ non-trivial eigenvectors (excluding the smallest eigenvalue $\lambda_0=0$). These spectral encodings act as low-frequency structural coordinates, enabling the GAT to distinguish between nodes with identical local neighborhoods but distinct global roles, such as bridge nodes versus peripheral nodes.

\item \textbf{Graph Attention Backbone:} The model consists of 4 GAT layers with multi-head attention ($M=4$ heads). For a node $i$, the output of a GAT layer is:
\begin{equation}
\mathbf{x}'_i = \sigma \left( \sum_{j \in \mathcal{N}(i)} \alpha_{ij} \mathbf{W}\mathbf{x}_j \right)
\end{equation}
where $\alpha_{ij}$ are the learnable attention coefficients computed via a shared self-attention mechanism on the node features. Each layer utilizes a hidden dimensionality of 256 ($64 \times 4$ heads), followed by Batch Normalization, ReLU activation, and Dropout ($p=0.3$).

\item \textbf{Residual Connectivity:} To facilitate the training of a deep graph architecture and preserve signal from the temporal inputs, we implement additive skip connections. For each layer $l$, the update rule is $\mathbf{x}^{(l)} = \text{ReLU}(\text{GAT}(\mathbf{x}^{(l-1)})) + \mathbf{x}^{(l-1)}$, ensuring that the mechanistic signal from the infection times is propagated through the network.

\item \textbf{Output Head and Inference:} The final node embeddings are passed through a linear layer to produce a scalar logit for each node. These are reshaped into a probability distribution over the $N=1000$ nodes in the graph using a Softmax activation. The model is trained using Cross-Entropy loss to maximize the likelihood assigned to the true source node.
\end{itemize}

Each node’s input consists of its observed infection time (or a sentinel value if unobserved) concatenated with its Laplacian positional encoding (LapPE) \cite{laplacianpe}, derived from the first 16 nontrivial eigenvectors of the graph Laplacian. These encodings provide global structural context and allow the model to distinguish between nodes with identical local neighborhoods but different roles in the global network topology—critical for inferring cascade origins. The GAT was trained to output a per-node probability distribution over possible sources, using cross-entropy loss.

\paragraph{Chemical Yield Prediction.}
For regression on simulated chemical reaction yields, we implemented a hybrid wide-and-deep SGNN architecture \cite{cheng2016widedeeplearning} specifically tailored to capture the discrete combinatorial nature of organic synthesis. This architecture simultaneously models global additive trends and local, high-order mechanistic interactions between reagents.

\begin{itemize}
\item \textbf{Inputs and Embedding Layer:} Each reaction is defined by five categorical variables: \textit{aryl halide}, \textit{boronate}, \textit{ligand}, \textit{base}, and \textit{solvent}. Each component is mapped to a 128-dimensional learnable embedding. These are concatenated to form a 640-dimensional base feature vector $\mathbf{x}_{\text{emb}}$.

\item \textbf{Wide Pathway:} A linear transformation maps $\mathbf{x}_{\text{emb}}$ directly to a scalar. This wide component acts as a generalized linear model, capturing the additive marginal effects of each reagent and providing a stable baseline for the predicted yield.

\item \textbf{Deep Pathway and Interaction Layer:} The deep component begins with a 128-unit "interaction" layer with ReLU activation that operates on the concatenated embeddings. The output is then concatenated with the original $\mathbf{x}_{\text{emb}}$ to form a 768-dimensional input for a multi-layer perceptron (MLP). The MLP consists of a series of dense layers (768 $\rightarrow$ 768 $\rightarrow$ 384 $\rightarrow$ 128 $\rightarrow$ 64) utilizing ReLU activations, Batch Normalization to stabilize the learning of synthetic distributions, and Dropout ($p \in [0.10, 0.15]$) to prevent the model from memorizing specific simulation artifacts.

\item \textbf{Multi-Head Output and Ensembling:} The final 64-dimensional latent representation is passed to three specialized heads:
    \begin{enumerate}
        \item \textit{Main Regression Head:} Predicts a yield value that is summed with the wide pathway output to produce the primary prediction $\hat{y}_m$.
        \item \textit{Auxiliary Head:} A 2-layer MLP (64 $\rightarrow$ 32 $\rightarrow$ 1) that produces a secondary yield estimate $\hat{y}_a$, encouraging the deep feature extractor to maintain a diverse representation of the reactivity space.
        \item \textit{Confidence Head:} An MLP (64 $\rightarrow$ 32 $\rightarrow$ 1) with a sigmoid activation that outputs a scalar $\hat{c} \in [0, 1]$. This head learns to weight the absolute residuals, effectively acting as a proxy for the model's epistemic uncertainty regarding specific reagent combinations.
    \end{enumerate}

\item \textbf{Dynamic Ensembling:} The final model output $\hat{y}_{\text{ens}}$ is a convex combination of the main and auxiliary heads: $\hat{y}_{\text{ens}} = w_m \hat{y}_m + w_a \hat{y}_a$. The weights $\mathbf{w}$ are derived from a two-element parameter vector passed through a softmax function, allowing the model to learn the optimal balance between the heads during pretraining.

\end{itemize}

This architecture is initially trained on the large synthetic corpus to internalize the underlying chemical rules. By leveraging the wide-and-deep structure, the SGNN learns a latent manifold of chemical reactivity that generalizes from structured simulations to sparse, noisy experimental data.

\subsection{Training Hyperparameters}

All models were trained using the AdamW optimizer \cite{adamw} with a weight decay of 0.01. Learning rates followed a cosine annealing schedule throughout training. Batch size was 256 for ecology, chemical yield prediction, and source identification, and 3,072 for disease forecasting (since the network was larger). The number of pretraining steps ranged from several hundred (ecology) to 10,000 steps (infectious disease forecasting). No fine-tuning was performed for any downstream evaluations; all results presented reflect direct zero-shot application of the pretrained SGNN models without additional task-specific retraining, except for in chemical yield prediction. Here, to demonstrate how fine-tuning could be applied, we pretrained on the simulated dataset and then fine-tuned on the real, smaller dataset \cite{perera2018platform}. Fine-tuning led to a 5\% increase in performance (fine-tuned performance shown in figure \ref{fig:general_results}b).

\subsection{Loss Functions and Optimization Objectives}

For each task, we used loss functions tailored to the specific requirements. For forecasting and regression, these objectives are designed to produce not only point estimates but also estimate the underlying uncertainty.

\paragraph{Probabilistic Forecasting and Parameter Inference.}
For time-series forecasting and the estimation of unobservable latent parameters (e.g., $R_0$), we used the quantile loss (also often called pinball loss). This objective allows the SGNN to produce a predictive distribution rather than a single value. For a target $y$ and a predicted quantile $\hat{y}_\tau$ at probability level $\tau \in (0, 1)$, the loss is defined as:

\begin{equation}
L_\tau(y, \hat{y}_\tau) = \max(\tau(y - \hat{y}_\tau), (1 - \tau)(\hat{y}_\tau - y))
\end{equation}

To generate calibrated 95\% prediction intervals, we optimize the model over a discrete set of $Q$ quantiles (e.g., $\tau \in \{0.025, 0.5, 0.975\}$). The total objective is the mean across all quantiles:
\begin{equation}
\mathcal{L}_{\text{forecast}} = \frac{1}{Q} \sum_{q=1}^{Q} L_{\tau_q}(y, \hat{y}_{\tau_q})
\end{equation}

\paragraph{Source Identification (Classification).}
We treat the identification of diffusion sources as a multi-class classification problem over $N$ nodes in the graph. The Graph Attention Network (GAT) outputs a probability distribution $\mathbf{\hat{y}}$ via a softmax activation. We minimize the cross-entropy loss:

\begin{equation}
\mathcal{L}_{\text{CE}} = -\sum_{i=1}^{N} y_i \log(\hat{y}_i)
\end{equation}

where $\mathbf{y}$ is the one-hot encoded ground-truth source vector. This objective encourages the model to assign maximum likelihood to the true mechanistic origin of the observed cascade.

\paragraph{Chemical Yield Regression (Multi-Head).}
For the Suzuki-Miyaura yield prediction, we employ a composite objective that balances the predictions of the hybrid wide-and-deep architecture with an uncertainty estimation head. The total loss $\mathcal{L}_{\text{chem}}$ is defined as:

\begin{equation}
\mathcal{L}_{\text{chem}} = \|y - \hat{y}_{\text{ens}}\|_2^2 + \alpha \|y - \hat{y}_a\|_2^2 + \beta \frac{1}{N} \sum_{i=1}^{N} \hat{c}_i |y_i - \hat{y}_{\text{ens}, i}|
\end{equation}

where $\hat{y}_{\text{ens}} = \sum_{k \in \{m, a\}} w_k \hat{y}_k$ is the weighted ensemble of the main ($m$) and auxiliary ($a$) heads, and $w_k$ are softmax-normalized learnable weights. The second term, scaled by $\alpha$, ensures predictive utility in the auxiliary head to promote representation diversity. The final term, scaled by $\beta$, incorporates the confidence head output $\hat{c} \in [0, 1]$, which acts as a learned weight on the absolute residuals to capture heteroscedastic noise and failure-prone regimes.

\section{Datasets and Evaluation}

We evaluated SGNNs across diverse forecasting, inference, and regression tasks using real-world datasets from epidemiology, chemical kinetics, and ecology.

\subsection{Infectious Disease Datasets and Evaluation}

\paragraph{COVID-19 Forecasting.}
We evaluated SGNNs on weekly COVID-19 mortality forecasting using data from the CDC COVID Forecast Hub \cite{cdccovid}, covering all 50 U.S. states, Washington D.C., and Puerto Rico. The evaluation period spanned April 2020 through November 2021, with weekly aggregation performed using CDC epidemiological weeks (epiweeks). SGNN models were applied zero-shot, without retraining or access to real COVID-19 data during training. Forecasting was conducted at horizons of 1 to 4 weeks ahead. Baseline models were evaluated over the same period, and all metrics were computed using the full real-world time series for each location. No normalization or transformations were applied.

\paragraph{Dengue Forecasting.}
Dengue forecasting was conducted using weekly case data from Brazilian states, sourced from Project Tycho \cite{tychodengue}. The SGNN was evaluated zero-shot, having been pretrained only on non-vector-borne simulations. Baseline models were trained directly on the dengue time series. For these baselines, we used a rolling-window evaluation strategy: models were retrained at each forecasting time point using all available data up to that point, to provide the strongest possible comparison. Forecast performance was then evaluated at 1–4 week horizons beyond each training window, and results were aggregated across all forecast periods and states.

\paragraph{$R_0$ Estimation.}
For synthetic evaluation of reproduction number ($R_0$) inference, we generated a held-out test set of 5,000 simulated epidemic trajectories using mechanistic models distinct from the pretraining corpus. Ground truth $R_0$ values were analytically computed via next-generation matrix methods, and estimation accuracy was evaluated using mean squared error and mean percentage error. For real-world validation, we evaluated SGNN and baseline methods on early COVID-19 case trajectories from March 2020 \cite{cdccovid}. Specifically, we used three weeks of daily reported cases per location (U.S. states, Washington D.C., and New York City) and compared SGNN $R_0$ estimates to values from the literature \cite{vissat2022comparison, ives2020state}. These real-world $R_0$ values were sourced from retrospective analyses using later data, providing a plausibility check against which early-estimate performance could be judged.

\paragraph{Back-to-Simulation Attribution.}
To evaluate SGNN interpretability, we used back-to-simulation attribution on real COVID-19 mortality time series in Michigan. The input sequence was passed through the SGNN encoder to extract its embedding. Cosine similarity was then computed between this embedding and the embeddings of 10,000 synthetic simulations used during pretraining. The top-50 most similar simulations were retrieved, and the distributions of their underlying mechanistic parameters were analyzed. This allowed us to examine whether SGNN's internal representation aligned real-world data with mechanistically plausible synthetic regimes.

\subsection{Ecological Datasets and Evaluation}

\paragraph{Lynx-Hare Forecasting.}
We evaluated SGNN on the canonical predator-prey dynamics of the Canadian lynx and snowshoe hare, using annual pelt-trapping records from 1821 to 1934 \cite{R_datasets_lynx}. The first 40 years of data were used for fitting mechanistic and statistical baselines and served as context input for SGNN. Models forecasted population counts 4 years ahead at each evaluation point. Evaluation metrics were averaged across forecast windows beyond the initial 40-year context.

\paragraph{Butterfly Population Forecasting.}
We evaluated SGNNs on high-dimensional ecological forecasting using the UK Butterfly Monitoring Scheme dataset \cite{UKBMS_2023}, which provides long-term annual abundance indices for dozens of butterfly species. Data through 2023 were obtained from the UKBMS public portal. Forecasting tasks were defined at multiple dimensionalities: for each value (e.g., 5, 10, 15 species), we selected species alphabetically (e.g., first 5, then next 5, etc.). Baseline models were trained directly on the real data, using the first 80\% of each species’ time series for training and the final 20\% for testing. SGNN was pretrained on synthetic simulations of multi-species population dynamics and applied zero-shot. No species from the real-world dataset were included in synthetic pretraining.

\subsection{Chemical Kinetics Dataset and Evaluation}

We evaluated SGNNs on reaction yield prediction for palladium-catalyzed Suzuki-Miyaura cross-coupling, using the nanomole-scale dataset developed by Perera et al.~\cite{perera2018platform} and accessed via the Open Reaction Database. The dataset consists of 5,760 real-world reactions screened under varying combinations of aryl halides, boronates, ligands, bases, and solvents, with yields reported as continuous values in the interval $[0, 1]$.

Each reaction was represented by five categorical input features corresponding to its component identities. No continuous descriptors (e.g., temperature or concentration) were used, and reactions were modeled purely based on discrete structural combinations. Reactions with missing yield values were excluded. The dataset was randomly split into 75\% for analysis (training + validation) and 25\% for held-out evaluation. All reported performance results use this final test set, which was never seen during training or pretraining.

After pretraining, the SGNN was evaluated on the held-out 20\% of real reactions. Model performance was assessed using the coefficient of determination ($R^2$), following the convention in prior work \cite{granda2018controlling}. The SGNN achieved a final $R^2$ of 0.90, outperforming previous state-of-the-art models trained directly on real data (which achieved $R^2 = 0.85$), corresponding to a one-third reduction in residual (unexplained) variance.

\section{Back-to-Simulation Attribution}

To enable mechanistically interpretable diagnostics, SGNNs support \textit{back-to-simulation attribution}, a retrieval-based method that identifies the most similar synthetic regimes to a given real-world input based on the SGNN's internal representation space. This section describes the implementation in more detail.

\paragraph{Embedding Extraction.}
To perform attribution, a real-world time series (e.g., weekly COVID-19 deaths in Michigan) is first transformed into the format expected by the pretrained SGNN encoder. This involves applying the exact same preprocessing used during model training. For example, in our case, log1p transformation followed by z-score normalization using training-set statistics. The model-specific context length is enforced, with earlier time points padded if necessary. The processed input is then passed through the full SGNN encoder, which, in this case, includes multi-scale convolutional feature extraction, categorical embeddings (e.g., disease type, time-series type, calendar indices), and a CNN-Transformer backbone. The encoder output is pooled using a validity mask over the time dimension to yield a fixed-length embedding $z \in \mathbb{R}^{d}$ (with $d = 1024$ in our primary implementation). While preprocessing steps and model architecture can vary by domain, attribution requires that embeddings be computed using the same pipeline and parameters as used during training.

\paragraph{Simulation Embedding Corpus.}
To enable attribution, we construct an embedding database by encoding a large set of synthetic trajectories from the SGNN pretraining corpus. Each simulation is processed using the same data pipeline as real-world inputs, including identical preprocessing steps (e.g., normalization, padding, feature construction) and the same trained SGNN encoder. The result is a set of embedding vectors $\{z_i\}$, each corresponding to a simulated trajectory. Alongside each embedding, we store the full set of underlying mechanistic parameters used to generate that simulation (e.g., transmission rates, population size, environmental forcing strength), which serve as the basis for interpretation. To support efficient reuse, all embeddings and parameter metadata are precomputed and cached to disk. While the specific preprocessing steps vary across domains, consistency between simulation and real-world input processing is essential for meaningful attribution.

\paragraph{Similarity Search.}
Given a real-world input embedding $z$, we compute cosine similarity against all stored simulation embeddings $\{z_i\}$:
\[
\text{sim}(z, z_i) = \frac{z \cdot z_i}{\|z\| \|z_i\|}.
\]
We retrieve the top-$k$ simulations with highest similarity scores (typically $k = 50$) using exact brute-force similarity via NumPy or scikit-learn. In practice, cosine similarity computation across 10,000+ embeddings completes in under 5 seconds on a single CPU core.

\paragraph{Mechanistic Interpretation.}
Each retrieved simulation is associated with its full parameter dictionary $\theta_i$. We analyze the top-$k$ simulations to inspect the distributions of their mechanistic parameters. For example, when applied to Michigan COVID-19 death data, the top retrieved simulations cluster tightly around known population size ($\sim$10 million), plausible hospitalization and death probabilities, and realistic asymptomatic proportions. This validates that the model's internal embedding space captures meaningful system-level dynamics and aligns real inputs with plausible mechanistic regimes.

\paragraph{Caching and Reproducibility.}
To accelerate repeated attribution queries, we cache both the simulation embeddings and parameter metadata as serialized binary files. These are loaded once per session and used for all future retrievals. If the cache does not exist, simulation embeddings are computed on demand and saved to disk.

\paragraph{Scalability.}
The full back-to-simulation attribution pipeline is implemented in PyTorch and NumPy and runs efficiently on CPU/GPU. Embedding a real input series requires only a forward pass through the encoder (under 100 ms on GPU), while similarity search over 10,000+ simulation embeddings takes under 5 seconds on CPU. This enables interactive diagnostic use cases and large-scale attribution across real-world datasets.

\section{Additional Results}\label{appendix:extended_data}

\subsection{Infectious Disease Forecast Evaluation}

To complement the forecasting skill analysis presented in the main text, we evaluated all models using standard probabilistic scoring metrics from infectious disease forecasting: the relative Weighted Interval Score (WIS) \cite{wis}, 50\% prediction interval coverage, and 95\% prediction interval coverage.

The Weighted Interval Score is a proper scoring rule that jointly evaluates the sharpness and calibration of probabilistic forecasts by averaging over a set of central prediction intervals \cite{wis}.
The WIS generalizes the absolute error to the probabilistic setting: it penalizes both wide intervals (poor sharpness) and observations falling outside predicted intervals (poor calibration), and reduces to the MAE when only a point forecast is provided.

To enable comparison across locations and time periods with different scales, we report relative WIS, computed as the ratio of each model's mean WIS to that of a persistence baseline. The persistence baseline forecasts the most recent observed value as its point prediction at all horizons, with prediction intervals derived from the historical variance of week-over-week changes in the target series. A relative WIS below 1.0 indicates improvement over the baseline, while values above 1.0 indicate worse performance. We additionally report empirical coverage of the 50\% and 95\% prediction intervals, where well-calibrated models should achieve coverage rates near these levels.

Results are presented in Table~\ref{tab:wis_covid} for COVID-19 mortality forecasting and Table~\ref{tab:wis_dengue} for dengue case forecasting. On COVID-19, SGNNs matched the relative WIS of the best CDC Forecast Hub model (0.63) while achieving better 50\% coverage (0.51 vs.\ 0.47). The PINN-based EINN model performed worse than the persistence baseline (relative WIS = 1.10) with poor calibration at both interval levels. On dengue, SGNNs again outperformed all baselines in relative WIS (0.80), with well-calibrated 95\% coverage (0.90). The ETS baseline achieved comparable 95\% coverage but with wider intervals reflected in its higher relative WIS (1.04).

\begin{table}[H]
\centering
\begin{tabular}{lccc}
\toprule
\textbf{Model} & \textbf{Relative WIS} $\downarrow$ & \textbf{50\% Coverage} & \textbf{95\% Coverage} \\
\midrule
SGNN & \textbf{0.63} & \textbf{0.51} & 0.91 \\
CDC Forecast Hub (Best) & \textbf{0.63} & 0.47 & \textbf{0.93} \\
Chronos & 0.88 & 0.42 & 0.82 \\
EINN (PINN) & 1.10 & 0.38 & 0.71 \\
\bottomrule
\end{tabular}
\caption{\textbf{Probabilistic forecast evaluation for COVID-19 mortality.} Relative WIS is computed against a persistence baseline (values $< 1.0$ indicate improvement). Nominal coverage targets are 0.50 and 0.95. Bold indicates best performance per column.}
\label{tab:wis_covid}
\end{table}

\begin{table}[H]
\centering
\begin{tabular}{lccc}
\toprule
\textbf{Model} & \textbf{Relative WIS} $\downarrow$ & \textbf{50\% Coverage} & \textbf{95\% Coverage} \\
\midrule
SGNN & \textbf{0.80} & \textbf{0.47} & \textbf{0.90} \\
ETS & 1.04 & 0.45 & \textbf{0.90} \\
EINN (PINN) & 2.03 & 0.31 & 0.64 \\
\bottomrule
\end{tabular}
\caption{\textbf{Probabilistic forecast evaluation for dengue case forecasting.} SGNNs were evaluated fully zero-shot with no exposure to dengue data or vector-borne transmission dynamics. Relative WIS is computed against a persistence baseline.}
\label{tab:wis_dengue}
\end{table}

\subsection{Ecological Forecasts}

\begin{figure}[H]
    \centering
    \includegraphics[width=0.95\textwidth]{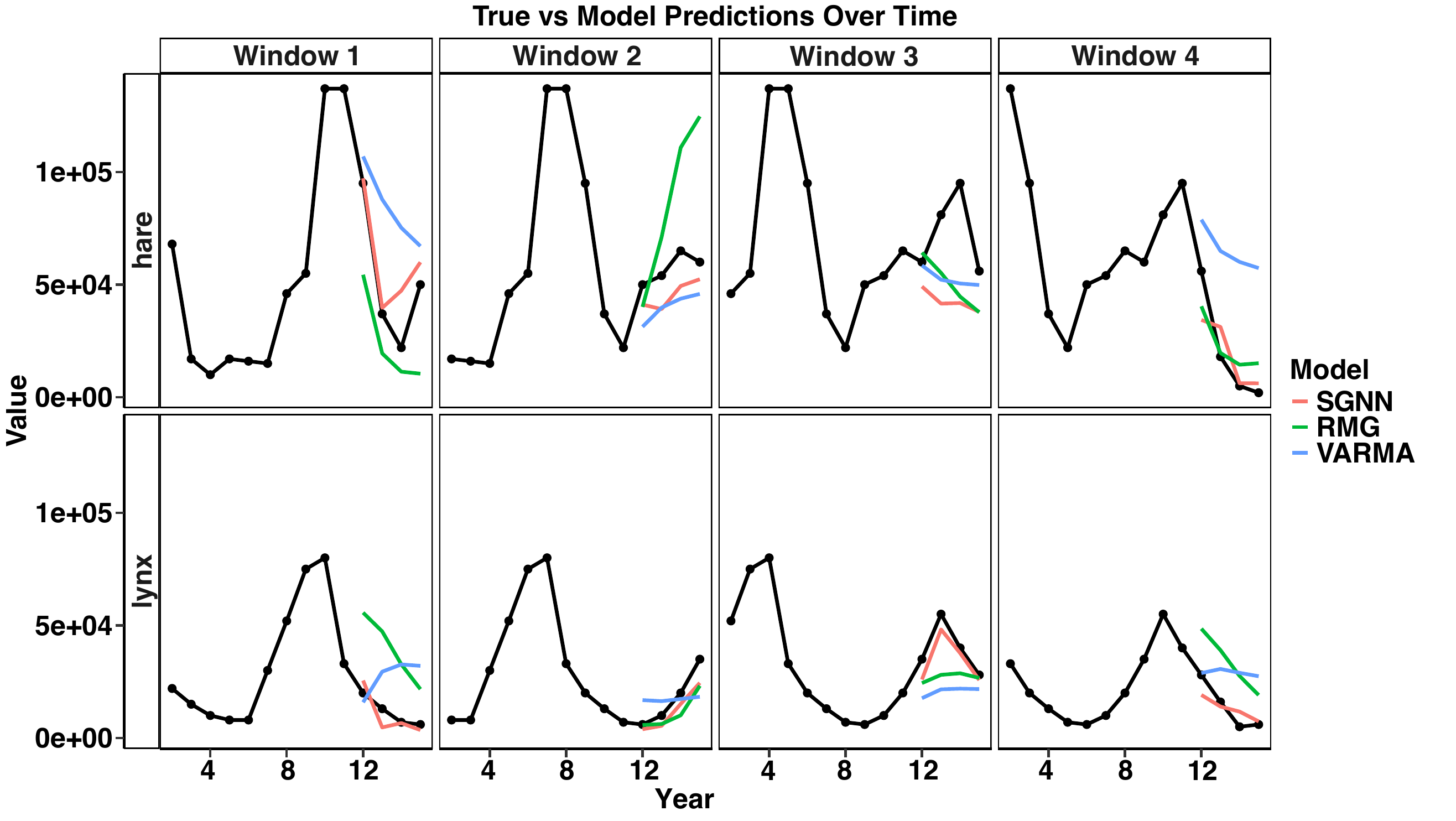}
    \caption{\textbf{SGNNs outperform classical models in forecasting lynx-hare predator-prey dynamics.} Rolling forecasts are shown for four evaluation windows, comparing SGNN (red) to mechanistic models (RMG, blue) and VARMA (green) models. True population trajectories are plotted in black. SGNNs maintain more accurate phase and amplitude tracking versus baselines.}
    \label{fig:lynxhare_appendix}
\end{figure}

To illustrate the forecasts of SGNNs in ecological systems, we provide a visual comparison of model predictions against ground-truth population dynamics in the lynx-hare predator-prey system. Figure~\ref{fig:lynxhare_appendix} shows forecast trajectories from SGNNs, Rosenzweig-MacArthur growth model (RMG), a purely mechanstic model, and vector autoregressive moving average models (VARMA), a purely data-driven model, across four rolling prediction windows.

\subsection{State $R_0$ Estimates}

To further evaluate SGNNs ability to infer unobservable epidemiological parameters, we estimated the basic reproduction number ($R_0$) for early COVID-19 outbreaks across all U.S. states and territories. Estimates were generated using SGNNs trained solely on synthetic data, and applied in a fully zero-shot setting to early outbreak curves (February–March 2020), consisting of three weeks of reported case data per location. Confidence intervals were taken from the quantile predictions of the SGNN.

SGNN-estimated early COVID-19 $R_0$ values across U.S. states and territories. Estimates reflect SGNN inference from early outbreak time series, using a model trained exclusively on synthetic simulation data. Confidence intervals denote 95\% quantile-based uncertainty.

\begin{minipage}{0.95\linewidth}
\begin{multicols}{2}
\setlength{\tabcolsep}{5pt}
\renewcommand{\arraystretch}{1.1}
\begin{tabular}{>{\bfseries}c@{\hspace{0.5em}}c}
Location & $R_0$ (95\% CI) \\
\hline
AK & 3.77 (3.19–6.06) \\
AL & 3.24 (2.08–3.80) \\
AR & 1.97 (1.53–2.52) \\
AS & 1.13 (0.90–1.27) \\
AZ & 2.09 (1.90–2.91) \\
CA & 4.95 (4.58–5.54) \\
CO & 1.46 (1.06–1.74) \\
CT & 1.10 (1.03–1.21) \\
DC & 1.87 (1.33–1.98) \\
DE & 1.59 (0.95–1.75) \\
FL & 1.68 (1.22–2.06) \\
FSM & 0.96 (0.80–1.27) \\
GA & 3.06 (2.72–3.13) \\
GU & 3.26 (2.96–5.63) \\
HI & 4.11 (3.03–5.37) \\
IA & 2.06 (1.45–2.90) \\
ID & 3.76 (2.43–5.80) \\
IL & 3.41 (2.86–3.55) \\
IN & 2.74 (1.97–3.00) \\
KS & 1.53 (1.17–1.71) \\
KY & 1.94 (1.45–2.49) \\
LA & 4.12 (3.86–5.31) \\
MA & 4.68 (4.05–5.69) \\
MD & 1.16 (0.82–1.73) \\
ME & 1.30 (1.28–1.73) \\
MI & 3.49 (3.09–3.67) \\
MN & 1.33 (1.24–2.18) \\
MO & 2.41 (1.86–3.11) \\
MP & 2.52 (1.73–4.21) \\
MS & 2.13 (1.64–2.47) \\
\end{tabular}

\columnbreak

\begin{tabular}{>{\bfseries}c@{\hspace{0.5em}}c}
Location & $R_0$ (95\% CI) \\
\hline
MT & 4.33 (3.98–6.08) \\
NC & 1.84 (1.37–1.92) \\
ND & 3.47 (2.74–5.26) \\
NE & 0.98 (0.85–1.49) \\
NH & 1.40 (1.38–2.32) \\
NJ & 3.02 (2.62–3.98) \\
NM & 0.75 (0.71–1.04) \\
NV & 4.08 (3.44–7.41) \\
NY & 6.11 (5.10–6.74) \\
NYC & 6.14 (5.02–7.97) \\
OH & 2.35 (2.11–2.17) \\
OK & 2.30 (1.64–2.77) \\
OR & 4.63 (4.25–9.04) \\
PA & 3.11 (2.80–3.13) \\
PR & 3.16 (2.07–4.52) \\
PW & 0.96 (0.80–1.27) \\
RI & 1.23 (1.07–1.57) \\
RMI & 0.96 (0.80–1.27) \\
SC & 1.80 (1.27–2.09) \\
SD & 0.89 (0.73–1.46) \\
TN & 2.87 (2.59–3.01) \\
TX & 2.98 (2.46–3.15) \\
UT & 1.80 (1.38–1.95) \\
VA & 0.97 (0.81–1.14) \\
VI & 6.79 (4.97–9.49) \\
VT & 2.91 (1.98–4.03) \\
WA & 4.80 (4.25–5.00) \\
WI & 1.59 (1.45–1.91) \\
WV & 2.20 (1.62–2.79) \\
WY & 2.66 (2.34–4.59) \\
\end{tabular}
\end{multicols}
\end{minipage}

\vspace{5mm}

The SGNN produced high and variable early-$R_0$ estimates across locations, with several important observations:

\begin{itemize}
    \item \textbf{High-$R_0$ regions}: NYC, NY, MA, CA, LA, MT, VI, and WA showed early $R_0$ values exceeding 4. These are consistent with retrospective studies that found substantial underreporting and rapid early spread, particularly in NYC, where early estimates underestimated true transmission by 2-3$\times$ \cite{vissat2022comparison, ives2020state}.

    \item \textbf{Low-$R_0$ regions and territories}: FSM, PW, RMI, NM, NE, and VA all yielded $R_0$ values near or below 1. This likely reflects true containment in isolated areas (e.g., FSM, RMI) or data sparsity in early periods, which limits identifiability and pulls predictions toward lower priors.

    \item \textbf{General interpretability}: SGNN’s estimates tend to reflect known drivers of early outbreak heterogeneity (population density, travel exposure, and reporting infrastructure) despite having no access to these covariates, suggesting that the SGNN can capture these patterns just from time-series shape and signal-to-noise characteristics.
\end{itemize}

\end{appendices}

\end{document}